\documentclass[11pt]{article}

\usepackage[preprint]{acl}
\usepackage{times}
\usepackage{latexsym}
\usepackage[T1]{fontenc}
\usepackage[utf8]{inputenc}
\usepackage{microtype}
\usepackage{inconsolata}
\usepackage{graphicx}
\usepackage{booktabs}
\usepackage{amsmath}
\usepackage{amssymb}
\usepackage{array}
\usepackage{xcolor}
\usepackage{colortbl}
\usepackage{stfloats}
\usepackage{algorithm}
\usepackage{algpseudocode}
\usepackage{tcolorbox}
\tcbuselibrary{skins,breakable}

\definecolor{promptframe}{RGB}{130,160,145}
\definecolor{promptback}{RGB}{247,249,248}
\newtcolorbox{promptboxbody}[1]{
  enhanced,breakable,
  colback=promptback,colframe=promptframe,
  colbacktitle=promptframe,coltitle=white,
  fonttitle=\small\bfseries,title={#1},
  title after break={#1 (continued)},
  boxrule=0.8pt,arc=2mm,
  left=6pt,right=6pt,top=6pt,bottom=6pt,
  before skip=6pt,after skip=8pt,
  fontupper=\small,
  before upper={\raggedright\setlength{\parindent}{0pt}}
}
\newenvironment{promptbox}[1]
  {\par\begingroup\nolinenumbers\begin{promptboxbody}{#1}}
  {\end{promptboxbody}\endgroup}

\setcitestyle{citesep={,}}

\newcommand{\arena}{Jev-ToM}

\title{Jev-ToM: Evaluating Theory of Mind in Typed Decision Models}
\author{
  Kawai Chung \quad Chunkit Chan \quad Yuxuan Liu \quad Yauwai Yim \\
  \textbf{Xi Yang \quad Wenbin Hu \quad Huihao Jing \quad Chiu Yeung Tang} \\
  \textbf{Changxuan Fan \quad Ki Sen Hung \quad Tsun On Kwok \quad Haoran Li} \\
  Department of Computer Science and Engineering \\
  The Hong Kong University of Science and Technology \\
  \href{https://github.com/HKUST-KnowComp/Jev-ToM}{\raisebox{-0.4pt}{\includegraphics[height=1em]{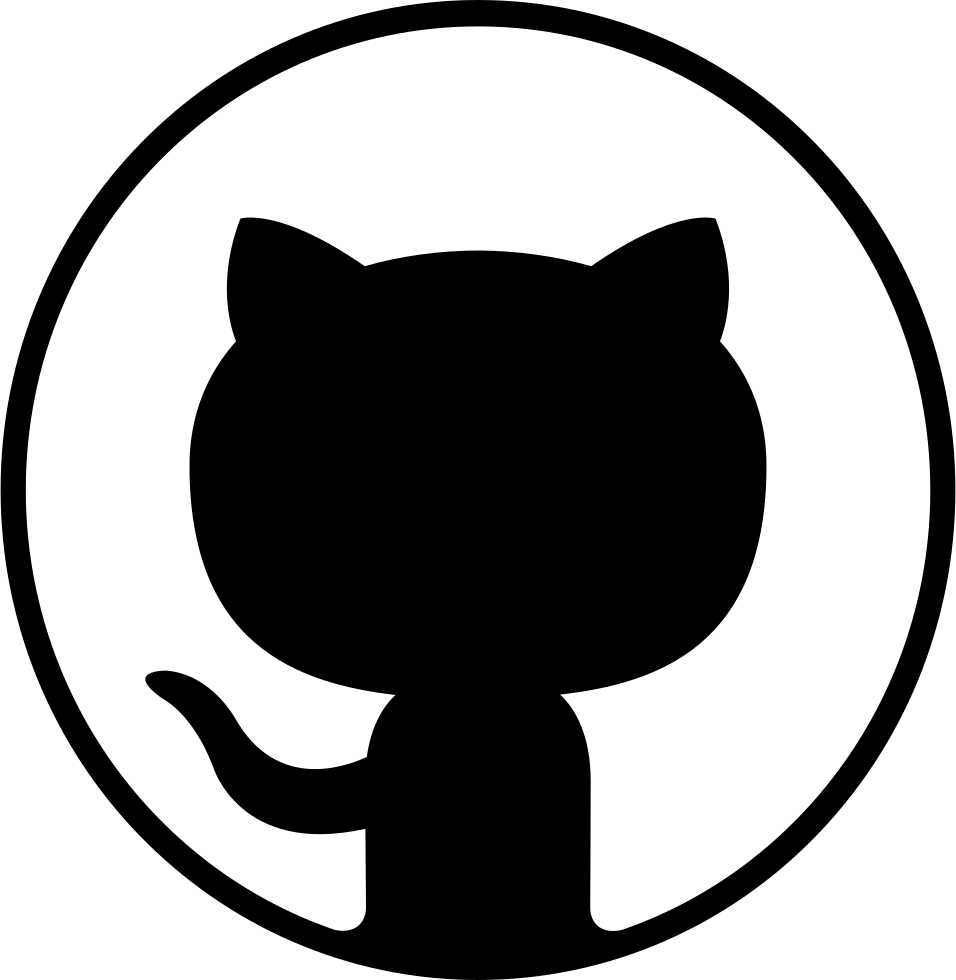}}~Code}\quad
  \texttt{\{kwchungac,ckchancc,hlibt\}@cse.ust.hk}
}

\begin{document}
\maketitle

\begin{abstract}
Dialogue and collaborative assistants repeatedly judge what a partner knows, believes, or intends. Typed decision models offer a potential alternative to generative large language models for bounded Theory of Mind (ToM) selection. We present \arena, comparing five typed decision models and seven generative baselines on 11,117 English text questions from nine ToM benchmarks across six capability groups. TypeSafe Jev reaches 73.2\% source-balanced accuracy including native controls and ranks first or second across all six capabilities, leading beliefs, goals/desires/intentions, and emotions. Other typed models show distinct strengths across tasks. Separate resource comparisons suggest that Jev could reduce the cost and waiting time of repeated ToM selection. Confidence-based selection remains benchmark-dependent, and supplied reasons can repair errors or overturn correct answers depending on their attachment. Candidate expansion reveals distinct selection and output-validity problems. Related-question cues account for most of Jev's downstream gains, with task-dependent effects from predicted states. The findings motivate testing Jev as a ToM selection component in assistants, with task-specific confidence calibration, candidate design, and output validation as integration priorities.
\end{abstract}

\section{Introduction}

\begin{figure}[!htbp]
\centering
\includegraphics[width=\columnwidth]{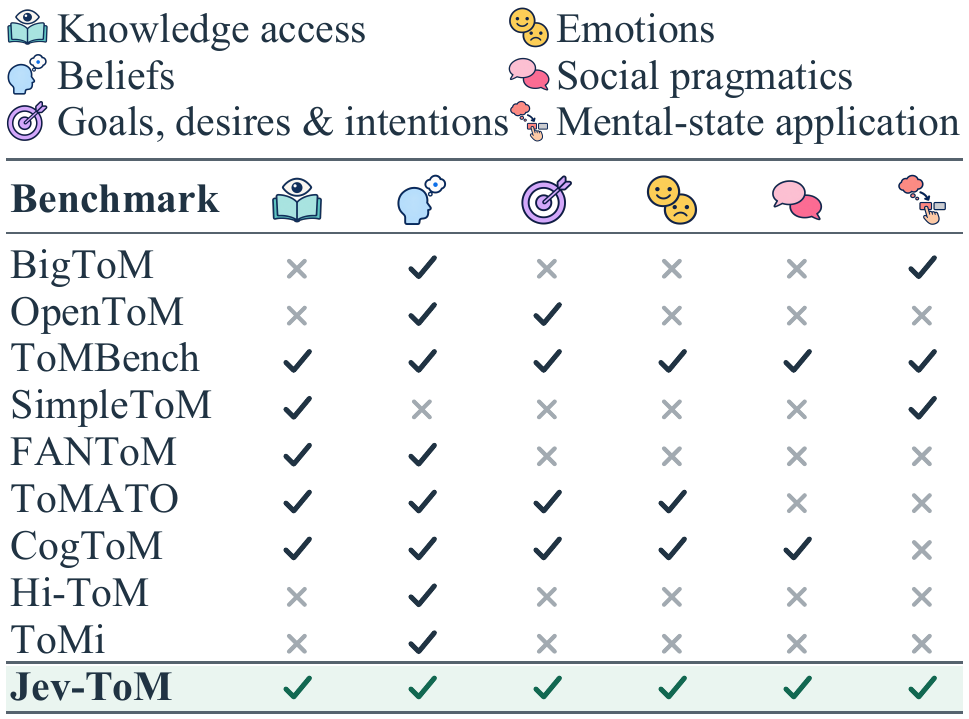}
\caption{Coverage of evaluated subsets and their union ($\checkmark$: present, $\times$: absent).}
\label{fig:capability_coverage}
\end{figure}

\raggedbottom
Theory of Mind (ToM) is the ability to attribute mental states, such as beliefs, knowledge, and intentions, to others and use them to interpret or predict behavior \citep{premack1978}. Advances in the language and reasoning capabilities of large language models (LLMs) have brought increasing attention to their ability to reason about others' perspectives \citep{strachan,xtom}. Useful interactive responses depend on the recipient's \mbox{perspective}, including missed information and outdated beliefs.

\begingroup
\widowpenalty=10000
Agents use mental-state inferences for action selection and response generation. Collaborative agents update explicit belief states from observations for planning \citep{tomcollab}. Dialogue agents use generative LLMs, including compact models, to infer participants' mental states from conversation and condition responses on those inferences \citep{toma}. For example, ToM-agent tracks a partner's beliefs, desires, and intentions in simulated persuasion dialogues, such as attempts to encourage a donation \citep{tomagent}. It maintains candidate interpretations, revises them using observed and predicted replies, and conditions its next response on the candidate with the highest confidence. The authors identify repeated LLM calls and long prompts as sources of waiting time and cost \citep{tomagent}. Repeated mental-state selection motivates testing Jev for candidate-based judgments in dialogue and collaboration, with candidate generation and response composition left to the surrounding system. Such intermediate judgments can have finite alternatives, even when final responses require open-ended language.\par
\endgroup

Typed decision models offer a direct interface for bounded judgments \citep{mnemon,jevbenchmark}. Given context, a question, and candidates, TypeSafe Jev (hereafter Jev) returns a selection and candidate probabilities as structured outputs \citep{jevdocs}. Its design targets fast, inexpensive decisions \citep{jevlaunch}. Structured selection makes Jev a promising candidate for bounded ToM judgments currently performed by generative LLMs. Fulfilling this role requires accurate mental-state judgments at low time and cost.

Fast, structured output alone does not establish suitability for this role. A high overall score can conceal differences between tracking knowledge, attributing beliefs or emotions, and applying mental states to behavior. Moreover, an agent needs to know when a selection is reliable and whether supplied reasons alter the choice or break the required output format. This raises a research question: \textbf{How competitive and reliable is Jev as a bounded ToM decision component?}

We present \arena, a systematic framework for evaluating Jev's ToM capabilities. To the best of our knowledge, this is the first dedicated evaluation of Jev's ToM capabilities across multiple benchmarks organized into explicitly defined capability groups. We compare five typed decision models and seven generative baselines on 11,117 questions from nine existing ToM benchmarks across six operational capability groups, as shown in Figure~\ref{fig:capability_coverage}. We examine confidence-based answer selection, interference from supplied reasons, and output validity, alongside observed response time and reference-price cost.

\begingroup
\widowpenalty=10000
Our results show that competitive accuracy can coexist with benchmark-dependent confidence errors, while supplied reasons can improve or disrupt selections depending on their attachment and the evaluated materials. The evaluation identifies confidence calibration, candidate sensitivity, and output validity as priorities for integration testing. Our contributions are threefold:\par
\endgroup

\pagebreak[0]
\begin{samepage}
\begin{list}{\textbullet}{\setlength{\leftmargin}{1em}\setlength{\labelsep}{0.5em}\setlength{\itemsep}{0pt}\setlength{\parsep}{0pt}\setlength{\topsep}{0pt}\setlength{\partopsep}{0pt}}
\item We compare Jev and generative LLMs across ToM tasks, measuring accuracy, reference cost, and observed response time.
\item We diagnose Jev's reliability through confidence-based selection and candidate interventions, separating selection errors from output-validity failures obscured by aggregate accuracy.
\item We investigate supplied reasons and mental-state feedback, testing reason attachment and separating the contributions of related-question cues and predicted states.
\end{list}
\end{samepage}
\newpage
\flushbottom
\begin{table}[!t]
\centering
\begingroup
\small
\setlength{\tabcolsep}{3pt}
\renewcommand{\arraystretch}{0.96}
\begin{tabular*}{\columnwidth}{@{\extracolsep{\fill}}lrrr@{}}
\toprule
Benchmark & Questions & Families & Options \\
\midrule
BigToM & 1,200 & 50 & 2 \\
OpenToM & 2,533 & 149 & 2/3 \\
ToMBench & 730 & 323 & 2/4 \\
SimpleToM & 891 & 177 & 2 \\
FANToM & 1,805 & 64 & 2 \\
ToMATO & 1,396 & 40 & 4 \\
CogToM & 2,154 & 1,215 & 4 \\
Hi-ToM & 154 & 30 & 15 \\
ToMi & 254 & 220 & 2 \\
\midrule
\textbf{Total} & \textbf{11,117} & \textbf{2,268} & 2/3/4/15 \\
\bottomrule
\end{tabular*}
\endgroup
\caption{Evaluated benchmark subsets.}
\label{tab:dataset_statistics}
\end{table}

\section{Related Work}

\paragraph{ToM agents and benchmarks.}
Mental-state judgments support cooperative planning \citep{tomcollab}, dialogue perspective tracking \citep{minddial,tomagent}, response selection \citep{toma}, and action prediction \citep{atom}. These uses motivate evaluating intermediate mental-state judgments in agents. ToM benchmarks differ in what successful judgment requires: controlled belief attribution and higher-order perspectives (beliefs about another person's beliefs) \citep{tomi,bigtom,hitom}, conversational information asymmetry \citep{fantom}, or diverse psychological states and social situations \citep{opentom,tombench,cogtom,tomato}. SimpleToM separates mental-state inference from behavior prediction and judgment \citep{simpletom}, while NegotiationToM studies inference in human negotiations \citep{negotiationtom}. This variety makes scope and aggregation choices central to interpreting a ToM score.

\paragraph{Jev interfaces and evaluation.}
Typed decision interfaces expose choices, scores, and probabilities for software to consume \citep{jevlaunch,jevdocs}. Mnemon uses Jev to evaluate retrieved records while LLM plans searches and composes responses \citep{mnemon}. This division motivates a similar component role for mental-state selection. Broad Jev evaluations already examine task performance, calibration, and resource use \citep{jevbenchmark}, alongside annotation and rubric assessment \citep{css,jevjudges}. Adversarial and option-definition studies establish that structured outputs can remain sensitive to request content \citep{jevadvbench,typesafe}. JevOut also includes ToMBench when testing contextual additions that redirect decisions \citep{jevout}. We extend these studies by connecting cross-benchmark ToM profiles with confidence-based selection, bidirectional changes under supplied reasons, and output validity.

\section{Experiments}
\label{sec:arena}

To assess breadth and reliability in bounded ToM selection, we compare performance on fixed answer choices and examine confidence, supplied reasons, and resource use in separate subsets of models and questions.

\subsection{Theory of Mind Ability Categories}
\label{sec:categories}

We adapt Abilities in Theory of Mind Space (ATOMS) to group questions by the mental states they concern \citep{beaudoin2020systematic}. \textbf{Knowledge access} covers information available through observation and communication \citep{fantom}. \textbf{Beliefs} concern representations of the world and others' perspectives \citep{bigtom,hitom}. \textbf{Goals, desires, and intentions} group related preferences and planned actions for cross-source comparison \citep{opentom}. \textbf{Emotions} covers feelings \citep{tomato}, while \textbf{social pragmatics} groups socially situated and non-literal meaning \citep{tombench}. CogToM's native Percept items remain in benchmark totals but outside capability groups pending finer mapping.

\textbf{Mental-state application} adds a different dimension: using a perspective to predict, select, or judge behavior \citep{simpletom}. A question asking what someone believes concerns attribution, whereas one asking how that belief affects their action concerns application. Application can overlap the state categories because it describes a task demand. Figure~\ref{fig:capability_coverage} also shows unequal source coverage: beliefs draw on eight benchmarks, while social pragmatics draws on two. These profiles compare models over the specified question groups.

\subsection{Dataset Selection and Harmonization}
\label{sec:datasets}

We assemble a common evaluation from nine existing ToM benchmarks with English contexts, reference answers, and discrete candidate choices \citep{bigtom,tomi,hitom,fantom,opentom,tombench,cogtom,tomato,simpletom}. Table~\ref{tab:dataset_statistics} summarizes the evaluated subsets. We use normal-length discrete OpenToM, short-context discrete FANToM, English ToMBench, and the ToMi subset accompanying SimToM \citep{simtom}. The collection consists of fixed development or previously used subsets, not newly collected questions. Appendix~\ref{app:statistics} gives their scope.

Source-specific adapters convert each question into a shared record containing source and question identifiers, a \emph{question family}, context, question, candidate texts, and a reference answer. Families group related questions or scenario variants using source identifiers or shared contexts, so a BigToM family can span multiple story variants. We retain the source contexts without truncation and preserve task-specific candidate meanings. Candidate counts remain unequal, ranging from two to fifteen. Harmonization therefore standardizes the evaluation record while retaining the original decision task.

We check that fields are nonempty, candidates are distinct, and references identify an offered answer. Of the initial 11,167 questions, we exclude 50 BigToM items for every model because their stories omit the subsequent event required by the reference. The resulting 11,117 questions and each question's candidate order are frozen across primary models. References are retained for scoring and withheld from model inputs. Source and family identifiers preserve the connection between standardized records and related judgments.

We then map native task, ability, or mental-state labels to the six capability groups. ToMBench application additionally follows the requested behavior prediction, selection, or judgment. All 730 ToMBench question targets underwent an agent-assisted post-evaluation review without item-level model predictions. This was not independent human validation. A question may enter multiple capability groups but counts once in its benchmark total. Appendix~\ref{app:mapping} provides the mapping rules.

\begin{figure*}[!t]
\centering
\includegraphics[width=0.96\textwidth]{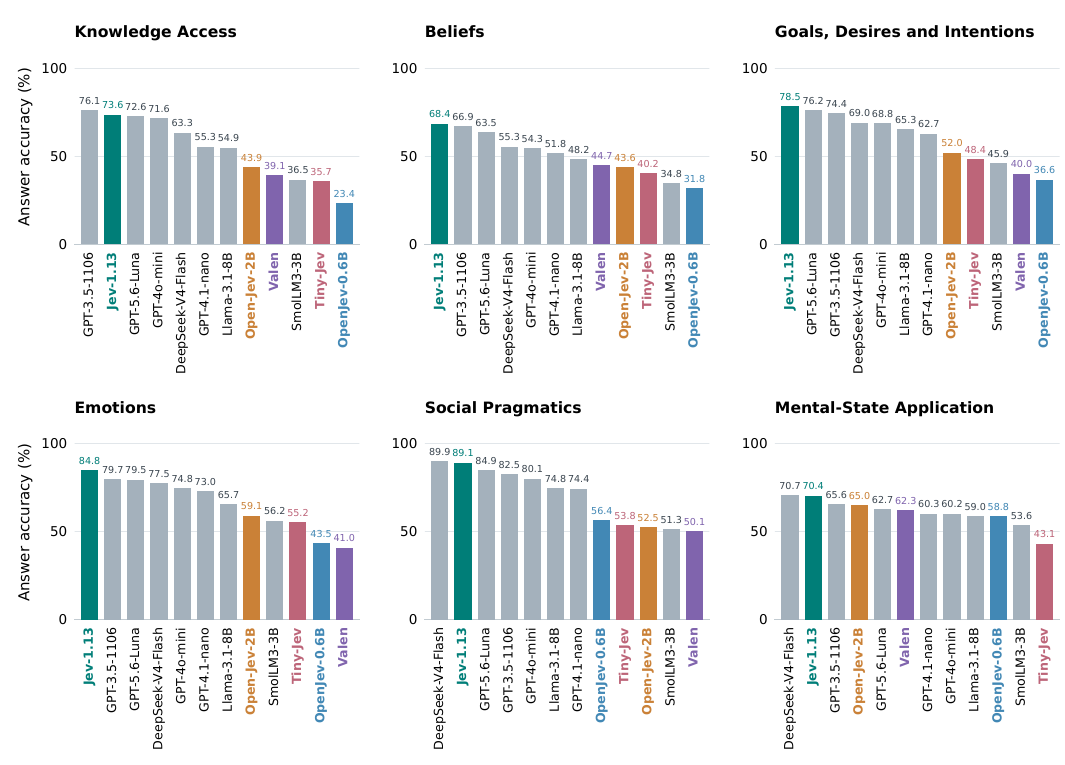}
\caption{Accuracy (\%) for knowledge access, beliefs, goals/desires/intentions, emotions, social pragmatics, and mental-state application, averaged equally across contributing benchmarks. Panels sort models by accuracy, with typed models colored and generative baselines gray. Benchmark control questions and unmapped questions are excluded.}
\label{fig:current_capabilities}
\end{figure*}

\subsection{Evaluation Metrics}
\label{sec:metrics}

Strict scoring requires a valid output and correct selection. Typed probabilities must cover exactly the offered choices, lie in $[0,1]$, sum to one, and give maximal probability to the selected choice, within numerical tolerances. Generated answers must resolve unambiguously to a candidate. Parsing and validation rules are detailed in Appendix~\ref{app:encoding}. Let $z_{mi}=1$ when model $m$ returns a valid answer matching the reference for item $i$, and $0$ otherwise, including invalid, unsupported, or failed responses. For benchmark $s$, let $D_s$ contain $N_s$ questions. Exact match (EM, \%) and source-balanced overall EM are
\begin{align}
\mathrm{EM}_{m,s} &= \frac{100}{N_s}\sum_{i\in D_s}z_{mi}, \\
\mathrm{EM}_{m,\mathrm{overall}} &= \frac{1}{9}\sum_{s=1}^{9}\mathrm{EM}_{m,s}.
\end{align}
For capability $c$, let $D_{s,c}$ contain its assigned questions from source $s$ and $\mathcal{S}_c$ the sources with nonempty $D_{s,c}$. We compute
\begin{equation}
\mathrm{EM}_{m,c}=\frac{1}{|\mathcal{S}_c|}\sum_{s\in\mathcal{S}_c}
\frac{100}{|D_{s,c}|}\sum_{i\in D_{s,c}}z_{mi}.
\end{equation}
Overall EM includes the benchmarks' original control questions and unmapped questions, while capability EM includes only assigned questions. An item may enter multiple capability groups but appears once per benchmark, so overall EM is not the mean of the six capability scores. Equal benchmark weighting prevents larger datasets from dominating. Pooled accuracy instead weights questions equally where explicitly stated. Further aggregation details are provided in Appendix~\ref{app:aggregation}.

\subsection{Evaluated Models}
\label{sec:model_selection}

We evaluated 12 models: five typed decision models and seven generative baselines. The typed models are TypeSafe Jev-1.13 \citep{jevlaunch}, Open-Jev-2B \citep{openjevcard}, Valen-Preview-0923 \citep{valencard}, Tiny-Jev \citep{tinyjevcard}, and OpenJev-0.6B \citep{openjev06card}. The generative baselines are SmolLM3-3B \citep{smollm3}, Llama-3.1-8B-Instruct \citep{llama31launch}, GPT-4.1-nano \citep{openai2025gpt41}, GPT-4o-mini \citep{openai2024gpt4omini}, GPT-3.5-Turbo-1106 \citep{openai2023devday}, GPT-5.6-Luna \citep{openai2026gpt56}, and DeepSeek-V4-Flash-0731 \citep{deepseekv4flash}.

\begingroup
\widowpenalty=10000
Primary models receive the same contexts, questions, and ordered candidates without reference labels. Typed interfaces encode the context as state and the question as instructions, returning a choice and candidate probabilities. Generative models receive the same content and return an option label. Both outputs are mapped back to the original candidates for strict scoring. Model-specific templates, decoding settings, and deployments are recorded in Appendix~\ref{app:settings}.\par
\endgroup

\section{Results and Analysis}
\label{sec:analysis}

\subsection{Where Are Typed Models Competitive?}
\label{sec:results}

\begingroup
\widowpenalty=10000
\paragraph{Jev is competitive across all six capabilities.}
Figure~\ref{fig:current_capabilities} ranks Jev first or second in every capability, with its largest lead over the strongest generative baseline in emotions: 84.8\% against GPT-3.5's 79.7\%. Strong relative rankings still leave substantial errors in beliefs and mental-state application, Jev's lowest-scoring capabilities. The other typed models remain below Jev but differ in their strengths. Open-Jev-2B leads this group in four capabilities, while Valen leads beliefs and OpenJev-0.6B social pragmatics. Tiny-Jev exceeds OpenJev-0.6B on emotions, reaching 55.2\% against 43.5\%, yet trails it on application. Application is the strongest capability for Open-Jev-2B, Valen, and OpenJev-0.6B, whereas Tiny-Jev favors emotions. Broad competitiveness is therefore a result for Jev, and the shared typed interface does not make the models interchangeable.\par
\endgroup

\begin{table*}[!t]
\centering
\begingroup
\fontsize{8}{9.2}\selectfont
\setlength{\tabcolsep}{2.1pt}
\renewcommand{\arraystretch}{1.04}
\begin{tabular*}{\textwidth}{@{\extracolsep{\fill}}l|*{10}{r}@{}}
\toprule
\textbf{Model} & \textbf{BigToM} & \textbf{OpenToM} & \textbf{ToMBench} & \textbf{SimpleToM} & \textbf{FANToM} & \textbf{ToMATO} & \textbf{CogToM} & \textbf{Hi-ToM} & \textbf{ToMi} & \textbf{Overall} \\
\midrule
\hyperlink{cite.jevlaunch}{\textbf{Jev-1.13}} & \cellcolor[HTML]{DDEBF7}\textbf{89.2} & \cellcolor[HTML]{DDEBF7}\textbf{60.0} & \cellcolor[HTML]{DDEBF7}\textbf{76.7} & \underline{59.1} & \cellcolor[HTML]{DDEBF7}\textbf{79.6} & \cellcolor[HTML]{DDEBF7}\textbf{83.6} & \cellcolor[HTML]{DDEBF7}\textbf{81.3} & \cellcolor[HTML]{DDEBF7}\textbf{59.7} & \underline{69.7} & \cellcolor[HTML]{DDEBF7}\textbf{73.2} \\
\hyperlink{cite.openjevcard}{\textbf{Open-Jev-2B}} & \underline{80.4} & \underline{45.2} & \underline{51.4} & 49.7 & \underline{71.5} & 50.1 & \underline{49.4} & \underline{39.6} & \cellcolor[HTML]{DDEBF7}\textbf{72.0} & \underline{56.6} \\
\hyperlink{cite.valencard}{\textbf{Valen}} & 74.9 & 38.1 & 46.0 & \cellcolor[HTML]{DDEBF7}\textbf{75.1} & 48.6 & 32.7 & 39.4 & 33.1 & 57.9 & 49.5 \\
\hyperlink{cite.tinyjevcard}{\textbf{Tiny-Jev}} & 73.9 & 34.0 & 41.5 & 20.9 & 71.2 & \underline{54.2} & 39.9 & 30.5 & 47.2 & 45.9 \\
\hyperlink{cite.openjev06card}{\textbf{OpenJev-0.6B}} & 74.6 & 39.1 & 40.7 & 56.9 & 29.1 & 20.0 & 42.5 & 31.8 & 53.9 & 43.2 \\
\midrule
\hyperlink{cite.openai2023devday}{\textbf{GPT-3.5-1106}} & \cellcolor[HTML]{E2EFDA}\textbf{86.9} & \cellcolor[HTML]{E2EFDA}\textbf{61.8} & \cellcolor[HTML]{E2EFDA}\textbf{75.1} & 55.2 & \cellcolor[HTML]{E2EFDA}\textbf{77.6} & \underline{81.6} & \underline{78.1} & \underline{55.8} & \cellcolor[HTML]{E2EFDA}\textbf{72.4} & \cellcolor[HTML]{E2EFDA}\textbf{71.6} \\
\hyperlink{cite.openai2026gpt56}{\textbf{GPT-5.6-Luna}} & 85.1 & 57.6 & \underline{73.8} & 49.7 & 72.2 & \cellcolor[HTML]{E2EFDA}\textbf{81.9} & \cellcolor[HTML]{E2EFDA}\textbf{81.5} & \cellcolor[HTML]{E2EFDA}\textbf{64.9} & \underline{70.9} & \underline{70.9} \\
\hyperlink{cite.openai2024gpt4omini}{\textbf{GPT-4o-mini}} & \underline{85.2} & 53.8 & 67.0 & 57.7 & \underline{75.3} & 75.1 & 72.2 & 42.9 & 67.3 & 66.3 \\
\hyperlink{cite.deepseekv4flash}{\textbf{DeepSeek-V4-Flash}} & 84.5 & \underline{60.2} & 70.0 & \cellcolor[HTML]{E2EFDA}\textbf{67.0} & 61.8 & 66.8 & 77.7 & 41.6 & 59.1 & 65.4 \\
\hyperlink{cite.openai2025gpt41}{\textbf{GPT-4.1-nano}} & 78.1 & 47.8 & 57.5 & 64.1 & 67.5 & 71.8 & 66.5 & 35.1 & 63.4 & 61.3 \\
\hyperlink{cite.llama31launch}{\textbf{Llama-3.1-8B}} & 76.7 & 48.4 & 59.9 & \underline{65.3} & 58.8 & 62.6 & 65.1 & 18.2 & 63.8 & 57.6 \\
\hyperlink{cite.smollm3}{\textbf{SmolLM3-3B}} & 78.3 & 45.0 & 48.1 & 47.7 & 42.1 & 35.2 & 49.7 & 35.1 & 57.1 & 48.7 \\
\bottomrule
\end{tabular*}
\endgroup
\captionof{table}{Exact-match accuracy (\%) on 11,117 questions, including native controls. Overall weights benchmarks equally. Invalid outputs count as errors. Bold/underlined values mark group best/second-best, with blue/green for typed/generative models. Excluding controls, Jev scores 69.1\% and GPT-3.5 70.9\%.}
\label{tab:benchmark_accuracy}

\end{table*}

\paragraph{Benchmark rankings can reverse.}
Table~\ref{tab:benchmark_accuracy} shows why overall rank is insufficient for model selection. Removing native controls reverses Jev's lead over GPT-3.5, leaving 69.1\% against 70.9\%. The comparison thus depends partly on which task demands enter the aggregate. Rankings also reverse within a specific benchmark: Open-Jev-2B reaches 72.0\% on ToMi against Jev's 69.7\%, despite trailing substantially overall. Appendices~\ref{app:benchmark_capabilities} and~\ref{app:control_sensitivity} provide capability breakdowns and control-free comparisons. Choosing a ToM component consequently requires matching its strengths to the intended tasks, rather than relying on a single aggregate score.

\begin{figure}[!ht]
\centering
\includegraphics[width=\columnwidth]{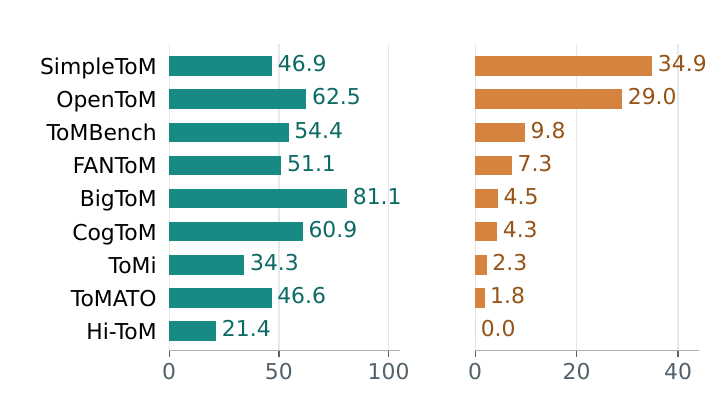}
\caption{Jev answer selection at confidence $\geq0.90$: coverage (left, \% of all questions) and retained error (right, \% of retained answers). Both panels follow descending error rate. Hi-ToM retains 33 answers.}
\label{fig:confidence_selection}
\end{figure}

\begin{figure*}[!t]

\centering
\includegraphics[width=0.90\textwidth]{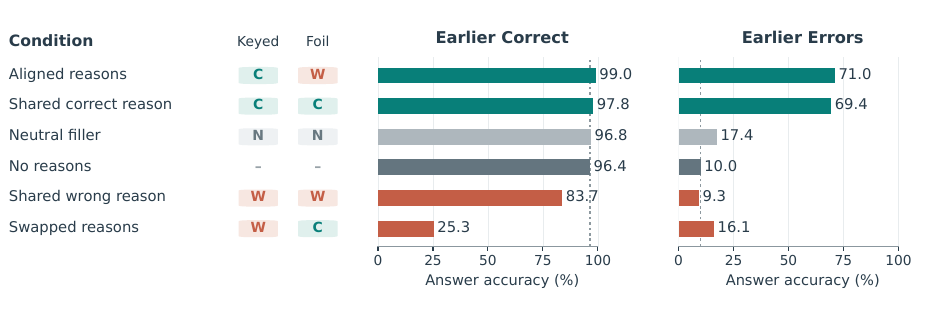}
\caption{Fresh Jev answer accuracy on earlier correct and incorrect cases. C/W/N/--: reference/selected incorrect/neutral/absent reasons. Dashed: fresh no-reason baseline.}
\label{fig:reason_controls}

\end{figure*}
\begin{figure*}[!b]

\centering
\includegraphics[width=\textwidth]{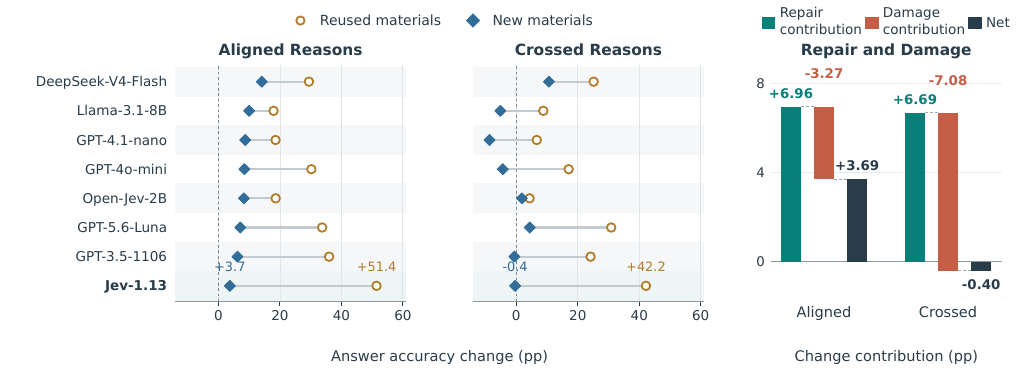}
\caption{Strict answer-component accuracy changes for aligned and crossed reasons, ordered by aligned gains on new materials. Repair and Damage: Jev's contributions (percentage points of all weighted outcomes) on new materials. Benchmarks receive equal weight.}
\label{fig:material_strata}

\end{figure*}
\subsection{Can Confidence Identify Reliable Answers?}
\label{sec:confidence}

\paragraph{Confidence supports selection unevenly across tasks.}
Figure~\ref{fig:confidence_selection} compares Jev's answer selection at a descriptive selected-answer probability cutoff of 0.90 across all 11,117 primary responses. \emph{Coverage} is the fraction retained and \emph{retained error} their error rate. Seven benchmarks fall below 10\% retained error, but SimpleToM and OpenToM remain at 34.9\% and 29.0\%. Similar coverage can therefore conceal very different reliability: ToMATO and SimpleToM both retain about 47\% of answers, yet their retained errors are 1.8\% and 34.9\%. Conversely, Hi-ToM's zero observed errors concern only 33 retained answers. Appendices~\ref{app:risk} and~\ref{app:risk_curves} provide counts and risk--coverage curves. The descriptive cutoff therefore needs task-specific validation before use as an acceptance policy.

\subsection{Intervention 1: How Do Supplied Reasons Affect Selection?}
\label{sec:reasons}
\label{sec:error_analysis}

\paragraph{Reason attachment can overturn an otherwise correct selection.}
We compare six conditions with questions, answer choices, and Jev settings fixed. The \emph{keyed} reason supports the reference answer, and the \emph{foil} reason supports a selected incorrect answer. \emph{Aligned} reasons accompany their own answers, \emph{swapped} reasons exchange these two attachments, and \emph{shared} conditions repeat either reason across answers. No-reason and neutral-text controls complete the comparison. Jev selects only an answer on 249 historically correct and 1,554 historically incorrect questions, each in original and reversed orders. Figure~\ref{fig:reason_controls} shows aligned reasons raising fresh accuracy on earlier-correct cases from 96.4\% to 99.0\%, while swapping lowers it to 25.3\%. Neutral text preserves 96.8\%, so extra text alone does not explain the loss. On earlier errors, aligned and shared keyed reasons reach 71.0\% and 69.4\% against a fresh 10.0\% baseline. Appendix~\ref{app:earlier_interventions} details sampling and conditions. Recovery can occur without option-specific support, yet misleading attachment can overturn a correct selection.

\paragraph{Small net changes conceal opposing outcome changes.}
In a separate eight-model study, \emph{materials} pair existing questions with supplied reasons. Three coauthors assessed reason support and attachment on an initial 450-question sample, with methods and majority-reference results in Appendix~\ref{app:human_annotation}. Models select bare answers or complete answer--reason pairs. Aligned pairs preserve each attachment, while \emph{crossing} includes all $K^2$ combinations of $K$ answers and reasons, including the keyed pair. Figure~\ref{fig:material_strata} compares the 1,803 preceding error-enriched materials, termed \emph{reused}, with \emph{new} reasons for 2,958 other existing questions. All gains use strict answer-component accuracy and equal benchmark weights. Aligned gains shrink on new materials for every model: Jev falls from 51.4 to 3.7 points, below Open-Jev-2B's 8.3-point gain. Crossed gains are lower still for all eight models and negative for five. Jev's Repair and Damage panel explains why the net effect is insufficient. Against matched no-reason answers, \emph{repair} changes errors to correct answers and \emph{damage} loses previously correct answers. From aligned to crossed, repair stays near seven points while damage rises from 3.3 to 7.1. Thus, 13.8\% of weighted outcomes change strict correctness despite a net $-0.4$ points. Crossing also increases candidate count and repetition. Appendix~\ref{app:fulloption} details material selection and repair--damage scoring. Reason sensitivity must therefore be assessed against both the material selection and the underlying changes in correctness.

\begin{figure}[!ht]
\centering
\includegraphics[width=\columnwidth]{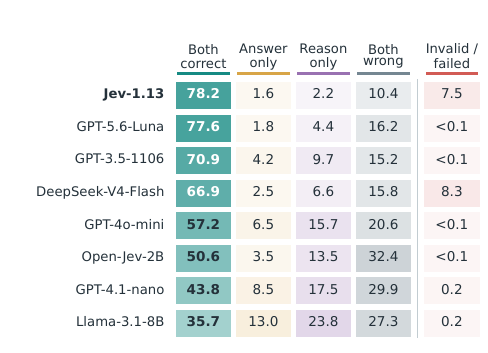}
\caption{Crossed-pair outcomes on all 4,761 materials, averaging both candidate orders and benchmarks equally. ``Both correct'' is strict joint accuracy, and ``only'' identifies the correct component. Failures stay in the denominator. Rows sort by both-correct share, with tints on a 0--100\% scale.}
\label{fig:joint_diagnostic}
\end{figure}

\begin{figure}[!ht]
\centering
\includegraphics[width=\columnwidth]{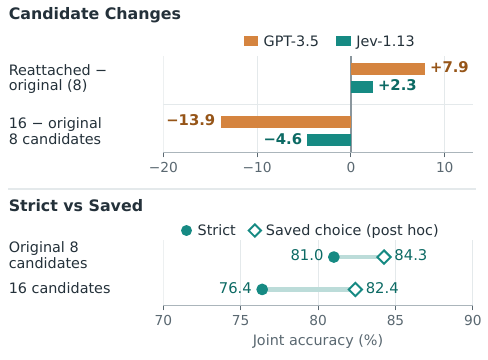}
\caption{Joint answer--reason accuracy on 108 questions (one per family) from CogToM, ToMATO, and ToMBench, with two orders and equal benchmark weights. Candidate Changes: strict changes from the original eight-pair set after reattachment or expansion to sixteen. Strict vs Saved: post-hoc scoring reads choices despite invalid probabilities without repairing them.}
\label{fig:matched_candidates_main}
\end{figure}
\paragraph{Keyed reasons can accompany wrong answers.}
Figure~\ref{fig:joint_diagnostic} separates answer/reason correctness from output validity in the same crossed responses, using benchmark-equal outcome shares. Strict \emph{joint accuracy} requires valid output with both components matching their keys. Every model selects only the keyed reason more often than only the keyed answer. For Llama, the shares are 23.8\% and 13.0\%, showing that selecting supplied support does not ensure a correct answer. Similar joint scores also conceal different failure profiles: Jev and Luna achieve 78.2\% and 77.6\% both-correct outcomes, but their invalid or failed shares are 7.5\% and less than 0.1\%, respectively. Open-Jev-2B rarely fails validation, yet its keyed-reason/wrong-answer share reaches 13.5\%. Appendix~\ref{app:output_failure_diagnostics} details probability \mbox{validation} and output diagnostics. Reliable pair selection requires checking semantic choices as well as the output contract.

\begin{figure*}[!htbp]

\centering
\includegraphics[width=\textwidth]{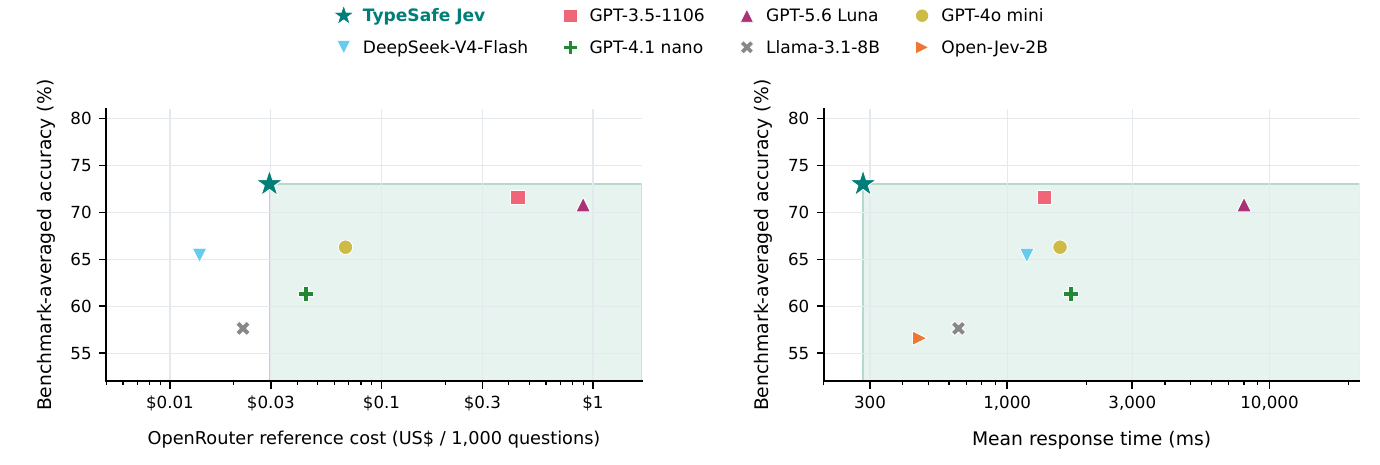}
\caption{Accuracy versus OpenRouter reference cost (left) and observed deployment time with differing concurrency (right). Shading marks lower accuracy with higher cost (left) or longer response time (right) than Jev.}
\label{fig:model_efficiency}

\end{figure*}

\paragraph{Candidate expansion affects both selection and output validity.}
To separate attachment from candidate expansion, we test 108 questions, one per family, using an eight-pair baseline that repeats four answers and four reasons twice. Reattachment changes only pairs with neither reference component. Expansion instead includes all sixteen combinations and changes repetition. Figure~\ref{fig:matched_candidates_main} shows reattachment helping both models, while expansion reduces accuracy, including a 13.9-point GPT-3.5 loss despite valid outputs. Jev's strict accuracy falls from 81.0\% to 76.4\%. Post-hoc \emph{saved-choice scoring} reads selections despite invalid probabilities without repairing them. Its smaller decline widens the strict-to-saved gap from 3.2 to 6.0 points, distinguishing selection loss from additional validation failures. Appendix~\ref{app:matched_candidates} gives the protocol. Even when the correct pair remains available, expanding the candidates can impair both selection and output validity.

\subsection{Intervention 2: Does Mental-State Feedback Improve Action Prediction?}
\label{sec:feedback_extension}

\begin{table}[!htbp]
\centering\small
\setlength{\tabcolsep}{3pt}
\begin{tabular*}{\columnwidth}{@{\extracolsep{\fill}}lrrrr@{}}
\toprule
Model / task & Original & \shortstack{Question\\only} & \shortstack{Predicted\\state} & \shortstack{Flipped\\state} \\
\midrule
Jev-1.13 & 78.50 & 89.25 & 93.00 & 86.25 \\
Open-Jev-2B & 66.50 & 64.00 & 64.75 & 61.75 \\
Valen & 62.75 & 60.50 & 59.75 & 60.00 \\
Tiny-Jev & 48.50 & 52.50 & 51.50 & 52.50 \\
OpenJev-0.6B & 61.50 & 55.00 & 51.25 & 51.00 \\
\midrule
Jev: BigToM & 92.00 & 98.00 & 97.50 & 97.00 \\
Jev: SimpleToM & 65.00 & 80.50 & 88.50 & 75.50 \\
\bottomrule
\end{tabular*}
\caption{Accuracy (\%) on 400 additional questions from families outside the primary sample, equally weighting 200 BigToM and 200 SimpleToM targets. The last rows show Jev by source. Its same-template fixed-No replay matches SimpleToM predicted feedback at 88.5\%.}
\label{tab:feedback_extension_main}
\vspace{-10pt}
\end{table}

\paragraph{Question cues account for most of Jev's gain.}
Table~\ref{tab:feedback_extension_main} compares five typed models on 400 behavior-prediction questions from families outside the primary sample. Each target shares a story with a belief/awareness question. \mbox{We compare} fresh original controls with \emph{question-only} cues supplying the related question, \emph{predicted-state} feedback adding the same model's answer, and \emph{flipped-state} feedback supplying the other binary answer, which need not be wrong. Question cues account for 10.75 of Jev's 14.5-point gain, with a further 3.75 points from adding the predicted answer. That extra gain concentrates on SimpleToM, where a post-hoc offline fixed-No replay of saved calls also reaches 88.5\% and reproduces every predicted-feedback selection. BigToM changes little across the three additions. Tiny-Jev favors question-only, while predicted feedback reduces accuracy for Open-Jev-2B, Valen, and OpenJev-0.6B relative to their original controls. Appendix~\ref{app:feedback_content_controls} details the replay and fixed-versus-predicted comparisons. Related-question cues can improve action selection, but the recorded gains do not establish a benefit from instance-specific mental-state predictions.

\subsection{What Are the Cost--Time Tradeoffs?}
\label{sec:efficiency}

\begingroup
\widowpenalty=10000
\paragraph{Jev improves the tradeoff against nano and mini.}
Figure~\ref{fig:model_efficiency} compares eight deployments on 11,117 questions, with seven priced using frozen OpenRouter reference tariffs. At US\$0.0295 per 1,000 questions, Jev costs 32.7\% less than nano and 56.3\% less than mini, with 11.7- and 6.7-point accuracy advantages. Lower cost alone does not favor Jev universally: DeepSeek offers US\$0.0138 at 65.4\% accuracy, giving Jev 7.6 more points at 2.14 times the cost. DeepSeek's longer observed response time also shows that lower cost need not mean less waiting.\newline
Jev's separate timing run records 73.0\% accuracy and a 282\,ms mean response time, the highest accuracy and shortest mean in this cohort. Its mean is 83.9\% below nano and 82.3\% below mini. Open-Jev-2B follows at 464\,ms with 56.6\% accuracy and appears only in the time comparison because it lacks a matching tariff. Appendix~\ref{app:efficiency} gives settings and full values. Observed timings reflect differing hardware, concurrency, and retry settings. Shorter decision calls could reduce waiting in assistants making repeated sequential judgments, but the overall benefit depends on the cost and time of generating candidates and composing responses, which this experiment does not measure.\par
\endgroup

\section{Conclusion}

\arena\ shows that Jev combines competitive bounded ToM selection with favorable reference-cost and observed-time comparisons in the evaluated deployments. Benchmark-dependent confidence errors and sensitivity to supplied reasons show why accuracy alone is insufficient for adoption. The results prioritize task-specific confidence calibration, candidate design, and output validation before testing Jev within complete assistants.

\section*{Limitations}

\paragraph{Language and task coverage.}
Our evaluation is limited to English, text-based questions with supplied candidates. The nine benchmarks cover several forms of ToM judgment, but their results do not establish generalization to other languages, cultural settings, or multimodal social situations. The six capability groups summarize the included tasks and are not an exhaustive measure of ToM. Possible training exposure to the public benchmarks also cannot be excluded.

\paragraph{Component-level evaluation.}
We evaluate Jev as a bounded decision component, without testing its integration into a complete interactive assistant. Candidate generation, partner-state updates across turns, and interactive task success are therefore outside our empirical claims. The confidence and reason analyses are exploratory, and the descriptive 0.90 cutoff is not a validated acceptance policy. The observed accuracy and resource advantages concern the evaluated tasks and deployments, so their value in a complete system depends on the surrounding components.

\section*{Ethical Considerations}

This study uses existing benchmark situations, model outputs, and annotations by three postgraduate students at English-medium universities with prior experience annotating ToM benchmarks. Annotation procedures are provided in Appendix~\ref{app:human_annotation}. It does not infer sensitive attributes of real users. Social-decision components can nevertheless harm users when predictions about beliefs or intentions are treated as facts. Applications should distinguish observed information from inferred states and validate the consequences of their errors. Our results do not justify autonomous consequential decisions about individuals. Dataset and model redistribution remains subject to source terms, with the planned release described in Appendix~\ref{app:release}.

\begingroup
\widowpenalty=10000
To support reproducibility, we plan to release a versioned package containing evaluation materials, subset and family identifiers, candidate-order mappings, per-question predictions and available probability vectors, de-identified annotation-stage records, and executable scoring and family-level resampling code. The package will cover the reason interventions and matched-candidate analyses, with source attribution and redistribution governed by the original licenses. Appendix~\ref{app:release} describes the planned release.\par
\endgroup

\begingroup
\widowpenalty=10000
We used GPT-6.1-Sol to generate and separately review the reason materials. These supplied reasons are experimental candidate text, not explanations produced by the evaluated models. Construction and prompts are documented in Appendices~\ref{app:pipeline} and~\ref{app:prompts}, respectively. Language-model assistance also supported the question-target mapping audit, analysis-code preparation, and manuscript revision. The authors remain responsible for the methods, claims, and final publication.\par
\endgroup

\bibliography{references}
\appendix
\section*{Appendix}

\makeatletter
\setlength{\@fptop}{0pt}
\setlength{\@fpsep}{14pt}
\setlength{\@fpbot}{0pt plus 1fil}
\setlength{\@dblfptop}{0pt}
\setlength{\@dblfpsep}{14pt}
\setlength{\@dblfpbot}{0pt plus 1fil}
\makeatother

\section{Extended Related Work}
\label{app:related_work}

\subsection{ToM benchmarks and application}
ToM evaluations differ in the evidence available to a character and the reasoning required of a model. ToMi controls answer-space regularities that allowed earlier tasks to be solved through dataset biases \citep{tomi}. BigToM uses causal templates for belief and action questions, while Hi-ToM extends perspective tracking to recursively nested beliefs \citep{bigtom,hitom}. Small alterations to otherwise similar tasks can also expose failures concealed by average success \citep{ullman}. Task structure and perspective depth therefore matter when interpreting a ToM score.

Broader benchmarks add psychological diversity and conversational information asymmetry. OpenToM incorporates personality, preferences, intentions, and physical and psychological states \citep{opentom}. ToMBench organizes diverse social-cognitive tasks and abilities, while CogToM draws on paradigms inspired by human cognition \citep{tombench,cogtom}. FANToM asks related questions about participants with unequal access to conversational information \citep{fantom}. ToMATO uses role-playing models' private thoughts to construct questions about beliefs, intentions, desires, emotions, and knowledge \citep{tomato}. NegotiationToM studies beliefs, desires, and intentions in human negotiations \citep{negotiationtom}. These constructions motivate reporting benchmark-specific results alongside capability groups.

SimpleToM distinguishes mental-state inference from behavior prediction and judgment, exposing a gap between explicit attribution and implicit application \citep{simpletom}. SimToM instead filters a story to information available to the queried character before answering a mental-state question \citep{simtom}. Our application category follows the requested judgment, and the feedback intervention separates the contribution of a related question from that of a predicted mental state.

\subsection{Jev evaluation and decision reliability}
TypeSafe introduces Jev as a model mapping unstructured state and typed questions to choices, scores, or probabilities \citep{jevlaunch,jevdocs}. Across 37 datasets, \citet{jevbenchmark} report strong results on many tasks but weaker performance for some label sets and judgment settings. The computational social science evaluation of \citet{css} finds substantial gaps from the best LLM annotators on many tasks despite lower measured costs. These findings favor evaluation of the intended decision domain and label space.

Confidence routing depends on both the task and the fallback. \citet{jevasjudge} test accepting confident Jev verdicts and escalating uncertain cases, with gains weakening under style-adversarial inputs and reference-free prose. In rubric judging, \citet{jevjudges} find shared errors between Jev and generative judges that limit accuracy gains from deferral. The annotation study likewise reports task-dependent confidence behavior \citep{css}. Our confidence analysis examines retained ToM answers without assuming that a fallback would correct them.

Well-formed outputs can still reflect unintended request cues. JevAdvBench compares single-edit attacks with clean and repeated calls \citep{jevadvbench}. \citet{typesafe} show that option-name polarity can redirect selections under unchanged definitions. JevOut optimizes natural-looking contextual additions that redirect initially correct decisions, including on ToMBench \citep{jevout}. Our predefined reason attachments probe repair and damage without adversarial context search. Probability Contracts distinguishes valid reports, coherent probabilities, and downstream decisions \citep{contracts}. An arithmetic rejection study further separates selecting an available answer from recognizing an answer-absent candidate set \citep{rejection}. Together, this work motivates checking semantic selection, numerical validity, and candidate coverage separately.

JEVal extends general decision-model evaluation to bilingual domains, agent workflows, and social simulation, finding that fast local decisions can coexist with lower long-horizon success and aggregate prediction errors \citep{jeval}. It also introduces InnerJev through reasoning-to-readout self-distillation. PACT studies contrastive evidence, answer-code permutations, and calibration objectives for single-token decisions \citep{pact}. These methods illustrate different routes to a decision interface rather than a common architecture for our five typed models.

\subsection{Multilingual and multimodal reasoning}
XToM constructs matched ToMi, FANToM, and NegotiationToM instances in English, Chinese, German, French, and Japanese through translation and human refinement \citep{xtom}. It distinguishes factual understanding from mental-state inference and finds that strong multilingual language understanding does not ensure comparable ToM performance across languages. A separate Jev cultural-values audit reports response and confidence changes across persona and language conditions \citep{persona}. Both make language and social framing relevant evaluation dimensions beyond our English-only comparison.

MultivationBench studies motivations inferred from accumulated visual and textual narratives \citep{multivationbench}. Maslow's expanded hierarchy of needs and Reiss's basic desires support four multi-label tasks pairing theoretical definitions with context-specific motivation options. This design distinguishes recognizing a psychological category from grounding it in behavior as a story unfolds. Its results identify difficulties in fine-grained motivation selection and correct judgments across a narrative. It complements belief-centered evaluation through visual grounding, multi-label motivations, and evolving evidence, which our fixed text questions do not measure.

\subsection{Mental-state reasoning in agents}
Explicit belief representations support cooperative planning \citep{tomcollab}. MindDial tracks speaker beliefs, predicted listener beliefs, and their differences for common-ground alignment and negotiation \citep{minddial}. ToMA connects mental-state generation with dialogue lookahead and interactive goal achievement \citep{toma}. Mnemon assigns record-level judgments to Jev while an LLM plans retrieval and composes responses \citep{mnemon}. These systems motivate evaluating repeated selection as a component alongside the surrounding interaction, as discussed in Appendix~\ref{sec:discussion}.

\section{Evaluation Protocol}
\label{app:settings}
\label{app:review_methods}

Table~\ref{tab:review_experiment_overview} summarizes the model cohorts, questions, candidate constructions, and aggregation for each comparison. Variants reuse source questions, so six conditions in two orders produce twelve responses per question. Primary, reason, feedback, and resource comparisons retain their own frozen populations.

\subsection{Capability mapping}
\label{app:mapping}
Native task, ability, and mental-state labels determine the six capability assignments in Table~\ref{tab:taxonomy}. Application additionally requires predicting, selecting, or judging behavior, including communicative appropriateness. Belief attribution, emotional reactions, factual recall, and utterance-truth judgments alone are excluded from application. The 152 ToMBench application assignments overlap other capabilities. All 730 ToMBench targets underwent an agent-assisted post-evaluation review without item-level model predictions, rather than independent human validation. A question counts once in its benchmark total.

\subsection{Datasets and harmonization}
\label{app:statistics}
The common primary set contains 11,117 questions. We use normal-length discrete OpenToM, short-context discrete FANToM, English ToMBench, and the ToMi subset accompanying SimToM \citep{simtom}, with fixed development or previously used subsets. Source adapters preserve contexts, references, candidate meanings, identifiers, and question families. Contexts are not truncated, and candidate order is frozen across models. From 11,167 initial questions, we exclude 50 BigToM \texttt{percept-to-belief} true-belief items whose stories omit the subsequent event required by the reference. Table~\ref{tab:appendix_statistics} reports question, family, option, and context-length statistics.

\subsection{Input, validation, and aggregation}
\label{app:interfaces}
\label{app:encoding}
\label{app:aggregation}
Typed models receive the context as \texttt{state}, the question as \texttt{instructions}, and indexed candidate texts, returning a choice and probabilities. Generative models receive the same content and return a candidate label. Strict scoring counts failed or unparseable outputs as wrong. Jev probabilities must cover exactly the offered choices, be finite and lie in $[0,1]$, sum to one within $10^{-4}$, and assign the selected choice a probability within $10^{-6}$ of the maximum. Confidence selection uses the selected-answer probability. Exact instructions and an input--output example appear in Appendix~\ref{app:prompts}.

Primary overall accuracy averages nine benchmark scores equally, including native controls and unmapped questions. Capability scores average contributing benchmarks and exclude controls and unassigned items. In the eight-model reason comparison, we average orders within questions, questions within benchmarks, and benchmarks equally. The six-condition intervention instead pools questions separately within each historical-correctness group. These weighting schemes remain distinct.

\subsection{Models and settings}
Table~\ref{tab:settings_summary} records the twelve models and their primary settings. Models receive one source question per request without task-specific fine-tuning. The six-condition study uses Jev, the full pair study uses eight models, matched candidates use Jev and GPT-3.5, and feedback uses five typed models. Resource settings are documented separately in Appendix~\ref{app:efficiency}.

\subsection{Paired uncertainty estimates}
\label{app:uncertainty}
Within each benchmark, bootstrap sampling resamples source-defined families with replacement, retaining all questions, paired outcomes, and candidate orders together. Each draw recomputes the original statistic. We use 5,000 draws for primary, material, and feedback comparisons, 2,000 for the six-condition study, and 10,000 for matched candidates. Percentiles 2.5 and 97.5 define pointwise 95\% intervals. Comparators remain fixed across draws. The estimates describe family-sampling variation, without adjustment for multiple comparisons or comparator selection, and do not measure repeated-inference or deployment variability. Complete estimates remain in the reproduction materials.

\section{Capability and Benchmark Results}
\label{app:details}

\subsection{Capability breakdowns}
\label{app:benchmark_capabilities}
Tables~\ref{tab:current-capabilities-1} and~\ref{tab:current-capabilities-2} expand Figure~\ref{fig:current_capabilities} with benchmark--capability scores for all twelve models. Missing capability cells are omitted, while controls and unmapped questions enter only benchmark totals. This separates a model's overall rank from its performance on the intended mental-state task.

\subsection{Sensitivity to native controls}
\label{app:control_sensitivity}
Table~\ref{tab:control_sensitivity} removes 920 native controls: 600 BigToM, 132 FANToM, 30 Hi-ToM, and 158 ToMi questions. The remaining 10,197 questions retain all 505 unmapped CogToM items and equal benchmark weighting. Jev falls from 73.2\% to 69.1\%, while GPT-3.5 changes from 71.6\% to 70.9\%, reversing their order. Their paired difference is $-1.79$ points $[-3.48,-0.07]$. ToMi contributes strongly to this reversal, with 50.0\% for Jev and 83.3\% for GPT-3.5 on its 96 non-control belief questions. The estimates use 5,000 within-benchmark family draws with seed 2026100401. ToMi retains 89 of its 220 families, and other benchmarks retain all families. The aggregate ranking consequently depends on whether native controls are included.

\section{Confidence-based Selection}

\subsection{Threshold and counts}
\label{app:risk}
The descriptive threshold of 0.90 applies to the probability assigned to the selected answer on all 11,117 Jev primary responses. Coverage divides retained answers by all questions in a benchmark. Retained error divides incorrect retained answers by retained answers. Table~\ref{tab:risk} gives the underlying counts for Figure~\ref{fig:confidence_selection}, including 33 retained Hi-ToM answers with no observed errors. The threshold is not a validated deployment policy.

\subsection{Risk--coverage curves}
\label{app:risk_curves}
Figure~\ref{fig:appendix_jev_risk_coverage} extends the comparison across all observed thresholds. Equal probabilities are retained together, and empty selections have undefined error. OpenToM and SimpleToM have similar full-coverage error, 40.0\% and 40.9\%, but retaining about half their answers yields 23.3\% error at 48.8\% coverage and 35.3\% at 48.9\% coverage, respectively. Confidence therefore separates errors more effectively in the former comparison even though both retain substantial error.

\section{Reason Interference and Candidate Selection}

\subsection{Material construction}
\label{app:pipeline}
For the eight-model pair study, 4,761 materials comprise 1,803 reused questions with reasons and newly prepared reasons for 2,958 other existing questions. Generation and review use benchmark references to construct supporting reasons for original answers, including plausible but mistaken reasons for distractors. Review checks story support, queried perspective, attachment, and overt reference leakage. This selects experimental materials rather than establishing human correctness.

Generation and separate review use GPT-6.1-Sol with JSON output. Newly prepared materials specify temperature 0, at most $\max(1600,200K)$ generation tokens for $K$ options, and 1,600 review tokens. Only service failures permit repeat requests. The exact generation and review prompts appear in Appendix~\ref{app:reason_prompts}.

\subsection{Six-condition answer selection}
\label{app:earlier_interventions}
\label{app:reasons}
Figure~\ref{fig:reason_controls} uses the 1,803 retained reused materials, comprising 249 historically correct and 1,554 historically incorrect Jev answers. Material screening starts with 531 correct and 2,533 incorrect cases. Generation and initial screening retain 2,142 cases, and subsequent review retains 1,803. Historical correctness defines the groups, while fresh no-reason calls provide their baselines.

Each question retains its original answers under six conditions: no reasons, identical neutral filler, aligned reasons, swapped keyed/foil reasons, the keyed reason repeated on every answer, and the foil repeated on every answer. The foil supports Jev's earlier incorrect choice, or the first non-keyed option for earlier-correct questions. Swapping changes only the keyed and foil attachments. Original and reversed orders are pooled separately within each historical group. Instructions and neutral text appear in Appendix~\ref{app:reason_instructions}. Complete plotted values remain in the reproduction materials.

\subsection{Human annotation}
\label{app:human_annotation}
Three postgraduate coauthors at English-medium universities, with prior ToM annotation experience, annotated in English as research collaborators without additional annotation payment. Written English--Chinese instructions and a label guide supported the procedure. Figure~\ref{fig:human_annotation_interface} preserves the complete interface excerpts.

The initial sample contains 50 questions per benchmark, totaling 450 questions and 1,879 reasons, stratified by material origin and option count without model predictions. Each annotator initially assessed 270 questions. All three shared 90 questions and 374 reasons, while 180 further questions had two annotations and 180 had one. A later answer-only stage collected 450, 450, and 270 judgments without adding reason labels. A follow-up covered all 249 earlier-correct intervention questions with three human judgments each. Nineteen exact-material overlaps yield 680 unique questions across stages. Table~\ref{tab:human_annotation_coverage} summarizes coverage and majority outcomes.

Annotators first answered from the story, question, and ordered options, recording evidence before reasons were revealed. Later revisions were saved separately. References, model predictions, machine review, and material origin were hidden. Story-support labels were \texttt{SUPPORTED}, \texttt{CONTRADICTED}, \texttt{NOT\_ESTABLISHED}, or \texttt{UNCERTAIN}. Attachment labels were \texttt{YES}, \texttt{NO}, or \texttt{UNCERTAIN}, asking whether the reason, if true, supported its answer. Non-\texttt{SUPPORTED}/\texttt{YES} judgments required a note. Answer judgments also allowed \texttt{MULTIPLE}, \texttt{NONE}, or \texttt{UNCERTAIN}.

For the initial two-annotator comparison on 374 shared reasons, story-support agreement was 85.56\% (Cohen's $\kappa=0.768$) and attachment agreement 95.19\% ($\kappa=0.833$). The later answer-only stage had mean pairwise agreement of 82.47\% on 270 questions shared by all three (Fleiss' $\kappa=0.751$). On the 90 initially shared questions, all three endorsed the keyed reason and attachment on 75, while at least one endorsed an alternative answer--reason pair on 30. Agreement and endorsement do not establish correctness or complete adjudication.

The post-hoc follow-up analysis establishes an answer-category reference when at least two of three annotators choose the same single category. For the first two annotators it uses the revised answer after reason review, or the initial answer if unchanged. The third vote comes from the independent record saved before other judgments were shown. Non-single-category judgments remain outside this subset. Of 249 questions, 219 have majority references, including 211 unanimous cases, and all 219 match the benchmark categories. The remaining 30 stay in the annotation records.

Table~\ref{tab:human_majority_intervention} retains both candidate orders and strict scoring, yielding 438 responses per condition on the majority subset. Aligned reasons raise accuracy from 98.2\% to 100.0\%, while swapped reasons lower it to 24.9\%. The paired changes are $+1.8$ and $-73.3$ points, with family-bootstrap intervals $[0.5,3.6]$ and $[-77.6,-68.9]$. Reason attachment therefore remains consequential on questions whose answer categories receive human majority support. The detailed annotation joins and exploratory R/K/S screens remain in the reproduction materials.

\subsection{Material selection, repair, and damage}
\label{app:fulloption}
\label{app:material_sensitivity}
\label{app:repair_damage}
For original answers $C=(c_1,\ldots,c_K)$ and reasons $E=(e_1,\ldots,e_K)$, the pair study compares bare answers, aligned pairs $(c_i,e_i)$, and all $K^2$ crossed pairs. Each condition uses a fixed shuffled order and its reversal. Answer-component scoring checks the selected answer, while joint accuracy requires both keyed components and a valid output. Crossing changes length, repetition, and candidate count as well as attachment.

Table~\ref{tab:material_results_current} supplies absolute reused/new accuracies for Figure~\ref{fig:material_strata}, together with Jev's paired repair and damage contributions. Repair changes a wrong no-reason outcome to a correct one, while damage changes a correct outcome to a wrong one. Both use all weighted outcomes, so the net change combines repair and signed damage, rather than conditioning on earlier errors or successes. On new materials, Jev's aligned change is $+3.69$ points $[1.93,5.51]$, while its crossed change is $-0.40$ $[-3.03,2.17]$. Small observed changes remain descriptive, and an interval spanning zero does not establish equivalence.

\subsection{Matched candidates}
\label{app:matched_candidates}
We select 40 accepted four-answer questions per source from CogToM, ToMATO, and ToMBench, one per family. Twelve families form the pilot and 108 questions the confirmation set. Every condition uses a fixed shuffled order and its reversal with equal benchmark weights. Let $(i,j)$ pair answer $i$ and reason $j$, with $(0,0)$ the reference pair. The eight-candidate sets are
\begin{align*}
E_{\mathrm{orig}}=\{&(0,0),(0,1),(1,1),(1,2),\\
                  &(2,2),(2,3),(3,0),(3,3)\},\\
E_{\mathrm{reassign}}=\{&(0,0),(0,1),(1,2),(1,3),\\
                      &(2,1),(2,3),(3,0),(3,2)\}.
\end{align*}
Each answer and reason appears twice. Reattachment preserves the reference pair's position, answer sequence, total length, and candidates with a reference component. Expansion compares the original eight pairs with all sixteen combinations. Four-pair controls use the original diagonal or $\{(0,0),(1,2),(2,3),(3,1)\}$. Table~\ref{tab:candidate_validity_review} reports all six conditions. Saved-choice scoring reads choices despite invalid probabilities without repairing them and remains post-hoc. Jev's reattachment change is $+2.31$ points $[-4.17,8.80]$, and its expansion change is $-4.63$ $[-10.19,0.93]$. GPT-3.5's expansion change is $-13.89$ $[-19.91,-7.87]$.

\subsection{Semantic outcomes and output validity}
\label{sec:joint}
\label{app:output_failure_diagnostics}
The five categories in Figure~\ref{fig:joint_diagnostic} partition requests into both components correct, only the answer correct, only the reason correct, neither correct, and invalid/failed output. Table~\ref{tab:failure_summary} distinguishes probability failures from service failures on the full crossed cohort. Jev has 448 sum failures and three selected-maximum failures among 9,522 responses, while DeepSeek has 843 final HTTP 429 outcomes and four parsing failures. These pooled counts use a different denominator from the figure's benchmark-equal shares.

For example, one saved Jev response selects \texttt{c3} with probability $0.9400000000000001$, assigns $0.05$ to \texttt{c5}, and zero elsewhere. Its selection can be correct while its approximately $0.99$ sum fails the numerical contract. All 448 sum failures differ from one by approximately $0.01$. A saved-response replay finds no change through sum tolerance $10^{-3}$, whereas $10^{-2}$ admits most of these outputs. Strict results retain the original $10^{-4}/10^{-6}$ tolerances. Complete candidate-count strata, tolerance replays, and response examples remain in the reproduction materials.

\section{Mental-state Feedback}

\subsection{Sample and conditions}
\label{app:mental_state_feedback_extension}
\label{app:feedback_extension}
We sample 200 BigToM belief/action pairs and 200 SimpleToM awareness/behavior pairs from families outside the primary evaluation, without consulting references or predictions. BigToM uses 100 families with true- and false-belief variants, balanced for explicit initial-belief statements. SimpleToM uses 200 families, 20 per scenario, with one story each and no behavior-judgment questions.

The five typed models first answer the upstream question. Fresh \emph{Original} controls answer the unmodified downstream question. \emph{Question-only} adds the related question, \emph{Predicted-state} also adds the same model's selected answer text, and \emph{Flipped-state} supplies the alternative binary answer, which need not be wrong. Conditions preserve the story prefix, downstream question, candidates, and order, changing only the suffix. Appendix~\ref{app:feedback_prompts} gives its exact text. Table~\ref{tab:feedback_extension_tasks} reports all models and sources. BigToM and SimpleToM receive equal weight, and SimpleToM awareness references are uniformly No.

\subsection{Fixed-No content control}
\label{app:feedback_content_controls}
Table~\ref{tab:appendix_fixed_no} compares predicted feedback with a post-hoc fixed-No replay on all 200 SimpleToM targets. For each target, we select the archived predicted or flipped response whose supplied upstream answer is No, preserving the original attribution header and consulting no downstream correctness. Replayed inputs exactly match the saved originals. Jev obtains 177/200 correct answers, with every selected option identical to predicted feedback. A constant cue thus reproduces Jev's recorded selections on this subset. The predicted-minus-question-only change over both sources is $+3.75$ points $[1.75,6.00]$, but this gain does not establish a benefit from instance-specific mental-state predictions. Full upstream-correctness strata remain in the reproduction materials.

\section{Cost and Response Time}
\label{app:efficiency}
\label{app:newprotocol}
Table~\ref{tab:resource_timing_current} joins the values in Figure~\ref{fig:model_efficiency} with serving settings for 11,117 questions per deployment. Reference costs apply OpenRouter public input/output tariffs frozen on 1 October 2026, without batch or cache discounts, to all recorded attempts and request overhead. Totals include conservative allowances for missing token usage. Seven deployments have tariffs, with GPT-3.5 using a historical listing. Open-Jev-2B appears only in the timing panel. These estimates are reference prices rather than actual bills.

Mean response time runs from evaluation-request start to final outcome, including communication and transport retries but excluding queuing and model initialization. Jev and Open-Jev-2B have separate serial timing passes with one attempt per question. Other deployments use concurrent requests. Open-Jev-2B uses a local M4 Pro with 48 GiB, macOS 15.7.7, vllm-jev 0.1.0, PyTorch 2.13.0, and Transformers 5.17.0. Llama and DeepSeek use DeepInfra with provider fallback disabled. Direct generative calls use temperature 0 and 32 output tokens. Luna uses low effort in new tool-disabled sessions without that hard cap. Its recorded usage includes additional session context whose complete text was not archived. Provider rate limits and hosted hardware were not recorded. Exact timeouts, attempt limits, output caps, tariffs, and missing-usage accounting remain in the reproduction materials. These serving conditions distinguish the observed deployment comparison from isolated architectural latency.

\section{Potential of Jev Application}
\label{sec:discussion}
The capability profiles and resource comparisons motivate testing Jev as a selector in dialogue and collaborative agents \citep{minddial,tomcollab}. An LLM could propose partner-state interpretations from dialogue history and generate responses while Jev selects among them \citep{treeofthoughts}. Emotions and social pragmatics are promising starting points given Jev's observed domain scores, while belief attribution and application leave more room for improvement.

Candidates should distinguish the partner's perspective from the agent's information and be revised as evidence changes. Evaluate candidate coverage separately from selection accuracy when a correct interpretation is present. Reason interference motivates checking attachment and candidate sensitivity, while task-specific calibration can guide acceptance and clarification \citep{knowno}. Invalid outputs need a fallback. These component checks support a proposed integration study, with dialogue quality and collaborative task success evaluated in the complete system.

\section{Data Access and Reproduction}
\label{app:release}
Source benchmarks and models retain their access terms. BigToM, ToMBench, FANToM, CogToM, and SimToM use MIT licences, Hi-ToM Apache-2.0, SimpleToM ODC-By, OpenToM CC BY-NC 4.0, and ToMATO the Meta Llama 3 Community License. We plan to release our added reason materials with attribution under CC BY 4.0.

The versioned reproduction materials identify source questions and families, subsets, material versions, candidate orders and mappings, separate reference keys, predictions, and available probability vectors. Annotation stages distinguish initial answers, revisions after reason review, and adjudication using de-identified annotator IDs. Scoring and family-resampling code will retain both orders together. Complete table values, uncertainty estimates, detailed annotation screens and joins, tolerance replays, response examples, and execution settings are preserved separately from the concise appendix.

The prepared anonymous review archive contains materials, parsed primary and feedback outcomes, and aggregate reason results. Per-question reason outputs and matched-candidate records are archived separately. Complete replay coverage and release of these supplementary materials remain planned. Aggregate appendix data do not substitute for those per-question records.

\clearpage
\begin{table*}[!p]
\section*{Supplementary Tables and Figures}
\label{app:tables}
\centering\small
\setlength{\tabcolsep}{4pt}
\begin{tabular*}{\textwidth}{@{\extracolsep{\fill}}>{\raggedright\arraybackslash}p{.19\textwidth}>{\raggedright\arraybackslash}p{.26\textwidth}>{\raggedright\arraybackslash}p{.31\textwidth}>{\raggedright\arraybackslash}p{.17\textwidth}@{}}
\toprule
Study and cohort & Sample & Target and candidate construction & Aggregation \\
\midrule
Primary / 12 models & 11,117 questions, nine benchmarks & $K$ original answers, answer accuracy & Equal benchmark weights \\
\addlinespace[2pt]
Confidence / Jev & Same 11,117 primary questions & Selected-answer probability, coverage and retained error & Within benchmark \\
\addlinespace[2pt]
Six conditions / Jev & 1,803 materials, split 249 / 1,554 by earlier correctness & $K$ answers with six reason conditions, two orders & Pooled within historical group \\
\addlinespace[2pt]
Reason pairs / 8 models & 4,761 materials: 1,803 reused and 2,958 new & $K$ answers, $K$ aligned pairs, $K^2$ crossed pairs, two orders & Equal benchmark weights \\
\addlinespace[2pt]
Matched candidates / 2 models & 108 questions, one per family, three benchmarks & Joint selection among 4, 8, or 16 pairs, two orders & Equal benchmark weights \\
\addlinespace[2pt]
Feedback / 5 typed models & 200 BigToM and 200 SimpleToM pairs, 300 families & Upstream binary state, downstream selection under four suffix conditions & Equal benchmark weights \\
\addlinespace[2pt]
Resources / 8 deployments & 11,117 questions per deployment & Original answer selection, reference cost and response time & Benchmark-equal accuracy, mean request time \\
\addlinespace[2pt]
\bottomrule
\end{tabular*}
\caption{Overview of the main comparisons. Condition and order variants reuse questions. $K$ denotes original answer count. Costs use seven priced deployments, and timing uses eight. The full pair study also records reason-only controls.}
\label{tab:review_experiment_overview}
\end{table*}

\begin{table*}[!p]
\centering\small
\setlength{\tabcolsep}{5pt}
\begin{tabular*}{\textwidth}{@{\extracolsep{\fill}}lrrrrrrrrrr@{}}
\toprule
Benchmark & K & B & G & E & S & A & C & U & $N$ & Families \\
\midrule
BigToM & -- & 400 & -- & -- & -- & 200 & 600 & -- & 1,200 & 50 \\
OpenToM & -- & 2384 & 149 & -- & -- & -- & -- & -- & 2,533 & 149 \\
ToMBench & 78 & 218 & 145 & 103 & 186 & 152 & -- & -- & 730 & 323 \\
SimpleToM & 297 & -- & -- & -- & -- & 594 & -- & -- & 891 & 177 \\
FANToM & 1446 & 227 & -- & -- & -- & -- & 132 & -- & 1,805 & 64 \\
ToMATO & 305 & 289 & 546 & 256 & -- & -- & -- & -- & 1,396 & 40 \\
CogToM & 212 & 440 & 469 & 331 & 197 & -- & -- & 505 & 2,154 & 1,215 \\
Hi-ToM & -- & 124 & -- & -- & -- & -- & 30 & -- & 154 & 30 \\
ToMi & -- & 96 & -- & -- & -- & -- & 158 & -- & 254 & 220 \\
\bottomrule
\end{tabular*}
\caption{Benchmark-to-capability question counts. K/B/G/E/S/A denote knowledge, beliefs, goals/desires/intentions, emotions, social pragmatics, and application. C/U denote native controls and unmapped CogToM items. The 152 ToMBench application assignments overlap other capabilities. A dash indicates no assigned questions.}
\label{tab:taxonomy}
\end{table*}

\begin{table*}[!p]
\centering\small
\setlength{\tabcolsep}{5pt}
\begin{tabular*}{\textwidth}{@{\extracolsep{\fill}}lrrrr@{}}
\toprule
Benchmark & Questions & Families & Options & Words: median \\
\midrule
BigToM & 1,200 & 50 & 2 & 69 \\
OpenToM & 2,533 & 149 & 2/3 & 196 \\
ToMBench & 730 & 323 & 2/4 & 61 \\
SimpleToM & 891 & 177 & 2 & 36 \\
FANToM & 1,805 & 64 & 2 & 566 \\
ToMATO & 1,396 & 40 & 4 & 539 \\
CogToM & 2,154 & 1,215 & 4 & 76 \\
Hi-ToM & 154 & 30 & 15 & 298 \\
ToMi & 254 & 220 & 2 & 59 \\
\midrule
Total & 11,117 & 2,268 & 2/3/4/15 & -- \\
\bottomrule
\end{tabular*}
\caption{Evaluated-subset statistics. Words are whitespace-delimited units in the original state, excluding questions and options.}
\label{tab:appendix_statistics}
\end{table*}

\begin{table*}[!p]
\centering\small
\setlength{\tabcolsep}{4pt}
\begin{tabular*}{\textwidth}{@{\extracolsep{\fill}}p{.15\textwidth}p{.33\textwidth}p{.47\textwidth}@{}}
\toprule
Model & Checkpoint or version & Evaluation settings \\
\midrule
Jev-1.13 & \path{typesafe/jev-1.13} & Native choice interface. \\
Open-Jev-2B & \path{ZefanCai/Open-Jev-2B} & Native choice interface and checkpoint settings. \\
Valen & \path{Valen-Team/Valen-Preview-0923} & Native choice interface and checkpoint settings. \\
Tiny-Jev & \path{lostargon/Tiny-Jev} & Native choice interface and checkpoint settings. \\
OpenJev-0.6B & \path{IamBusy/OpenJev-0.6B} & Native choice interface and checkpoint settings. \\
SmolLM3-3B & \path{HuggingFaceTB/SmolLM3-3B}, revision \texttt{a07cc9a04f16} & Non-thinking greedy decoding, 32 output tokens, batch size four, 16,384-token input limit. \\
GPT-4.1-nano & \path{gpt-4.1-nano-2025-04-14} & Temperature 0, 32 output tokens. \\
GPT-4o-mini & \path{gpt-4o-mini-2024-07-18} & Temperature 0, 32 output tokens. \\
GPT-3.5-1106 & \path{gpt-3.5-turbo-1106} & Temperature 0, 32 output tokens. \\
GPT-5.6-Luna & \path{gpt-5.6-luna} & Low effort, tool use disabled, no 32-token hard cap. \\
Llama-3.1-8B & \path{meta-llama/llama-3.1-8b-instruct} & Temperature 0, 32 output tokens. \\
DeepSeek-V4-Flash & \path{deepseek/deepseek-v4-flash-0731} & Non-thinking mode, temperature 0, 32 output tokens. \\
\bottomrule
\end{tabular*}
\caption{Model versions and primary evaluation settings. All models receive one question per request without task-specific fine-tuning.}
\label{tab:current_roster}
\label{tab:settings_summary}
\end{table*}

\begin{table*}[!p]
\centering\scriptsize
\setlength{\tabcolsep}{2.3pt}
\renewcommand{\arraystretch}{1.12}
\begin{tabular*}{\textwidth}{@{\extracolsep{\fill}}lrrrrrrrrrrrrr}
\toprule
Capability & $N$ & \shortstack{Jev\\1.13} & \shortstack{GPT-3.5\\1106} & \shortstack{GPT-5.6\\Luna} & \shortstack{GPT-4o\\mini} & \shortstack{DeepSeek\\V4-Flash} & \shortstack{GPT-4.1\\nano} & \shortstack{Llama-3.1\\8B} & \shortstack{Open-Jev\\2B} & \shortstack{Valen} & \shortstack{SmolLM3\\3B} & \shortstack{Tiny-Jev} & \shortstack{OpenJev\\0.6B} \\
\midrule
\multicolumn{14}{l}{\textbf{Cross-source capability atlas}} \\
Knowledge access & 2,338 & 73.6 & \textbf{76.1} & 72.6 & 71.6 & 63.3 & 55.3 & 54.9 & 43.9 & 39.1 & 36.5 & 35.7 & 23.4 \\
Beliefs & 4,178 & \textbf{68.4} & 66.9 & 63.5 & 54.3 & 55.3 & 51.8 & 48.2 & 43.6 & 44.7 & 34.8 & 40.2 & 31.8 \\
Goals, desires and intentions & 1,309 & \textbf{78.5} & 74.4 & 76.2 & 68.8 & 69.0 & 62.7 & 65.3 & 52.0 & 40.0 & 45.9 & 48.4 & 36.6 \\
Emotions & 690 & \textbf{84.8} & 79.7 & 79.5 & 74.8 & 77.5 & 73.0 & 65.7 & 59.1 & 41.0 & 56.2 & 55.2 & 43.5 \\
Social pragmatics & 383 & 89.1 & 82.5 & 84.9 & 80.1 & \textbf{89.9} & 74.4 & 74.8 & 52.5 & 50.1 & 51.3 & 53.8 & 56.4 \\
Mental-state application & 946 & 70.4 & 65.6 & 62.7 & 60.2 & \textbf{70.7} & 60.3 & 59.0 & 65.0 & 62.3 & 53.6 & 43.1 & 58.8 \\
\midrule
\multicolumn{14}{l}{\textbf{BigToM}} \\
Beliefs & 400 & \textbf{74.5} & 68.2 & 65.0 & 67.0 & 64.2 & 62.5 & 61.5 & 57.5 & 56.2 & 62.0 & 55.8 & 53.8 \\
Mental-state application & 200 & \textbf{93.5} & 93.0 & 92.0 & 84.5 & 87.5 & 60.0 & 63.0 & 76.5 & 62.0 & 60.5 & 59.5 & 53.5 \\
Control & 600 & 97.5 & 97.3 & 96.2 & \textbf{97.7} & 97.0 & 94.5 & 91.3 & 97.0 & 91.7 & 95.2 & 90.8 & 95.5 \\
All & 1,200 & \textbf{89.2} & 86.9 & 85.1 & 85.2 & 84.5 & 78.1 & 76.7 & 80.4 & 74.9 & 78.3 & 73.9 & 74.6 \\
\midrule
\multicolumn{14}{l}{\textbf{OpenToM}} \\
Beliefs & 2,384 & 60.1 & \textbf{62.5} & 57.9 & 54.1 & 61.0 & 48.2 & 48.6 & 45.6 & 38.2 & 45.2 & 33.1 & 39.1 \\
Goals, desires and intentions & 149 & \textbf{57.7} & 49.7 & 53.0 & 49.0 & 47.7 & 41.6 & 45.6 & 40.3 & 35.6 & 42.3 & 48.3 & 39.6 \\
All & 2,533 & 60.0 & \textbf{61.8} & 57.6 & 53.8 & 60.2 & 47.8 & 48.4 & 45.2 & 38.1 & 45.0 & 34.0 & 39.1 \\
\midrule
\multicolumn{14}{l}{\textbf{ToMBench}} \\
Knowledge access & 78 & 53.8 & \textbf{65.4} & 51.3 & 55.1 & 43.6 & 24.4 & 29.5 & 26.9 & 23.1 & 25.6 & 9.0 & 19.2 \\
Beliefs & 218 & 80.3 & \textbf{83.9} & 81.7 & 70.6 & 67.9 & 61.5 & 61.0 & 57.8 & 49.1 & 50.0 & 40.4 & 35.8 \\
Goals, desires and intentions & 145 & \textbf{75.9} & 71.7 & 73.1 & 63.4 & 70.3 & 57.2 & 65.5 & 45.5 & 38.6 & 43.4 & 35.2 & 35.2 \\
Emotions & 103 & \textbf{74.8} & 69.9 & 68.0 & 68.0 & 68.9 & 61.2 & 54.4 & 60.2 & 45.6 & 52.4 & 54.4 & 48.5 \\
Social pragmatics & 186 & \textbf{83.9} & 74.2 & 78.0 & 69.9 & \textbf{83.9} & 65.1 & 69.9 & 53.8 & 58.1 & 56.5 & 54.3 & 55.4 \\
Mental-state application & 152 & \textbf{73.7} & 65.8 & 65.1 & 55.3 & 69.1 & 57.2 & 57.2 & 53.3 & 42.1 & 46.1 & 39.5 & 39.5 \\
All & 730 & \textbf{76.7} & 75.1 & 73.8 & 67.0 & 70.0 & 57.5 & 59.9 & 51.4 & 46.0 & 48.1 & 41.5 & 40.7 \\
\midrule
\multicolumn{14}{l}{\textbf{SimpleToM}} \\
Knowledge access & 297 & 89.2 & 89.6 & 87.2 & \textbf{91.2} & 90.2 & 65.0 & 82.5 & 18.9 & 59.9 & 34.7 & 1.7 & 3.7 \\
Mental-state application & 594 & 44.1 & 38.0 & 31.0 & 40.9 & 55.4 & 63.6 & 56.7 & 65.2 & 82.7 & 54.2 & 30.5 & \textbf{83.5} \\
All & 891 & 59.1 & 55.2 & 49.7 & 57.7 & 67.0 & 64.1 & 65.3 & 49.7 & \textbf{75.1} & 47.7 & 20.9 & 56.9 \\
\bottomrule
\end{tabular*}
\caption{Capability exact match (\%), part 1. Cross-source rows average contributing benchmarks, while source rows pool assigned questions. Controls and unassigned items enter only All. Overlapping application assignments do not sum to benchmark totals. Bold marks the highest point estimate.}
\label{tab:current-capabilities-1}
\end{table*}

\begin{table*}[!p]
\centering\scriptsize
\setlength{\tabcolsep}{2.3pt}
\renewcommand{\arraystretch}{1.12}
\begin{tabular*}{\textwidth}{@{\extracolsep{\fill}}lrrrrrrrrrrrrr}
\toprule
Capability & $N$ & \shortstack{Jev\\1.13} & \shortstack{GPT-3.5\\1106} & \shortstack{GPT-5.6\\Luna} & \shortstack{GPT-4o\\mini} & \shortstack{DeepSeek\\V4-Flash} & \shortstack{GPT-4.1\\nano} & \shortstack{Llama-3.1\\8B} & \shortstack{Open-Jev\\2B} & \shortstack{Valen} & \shortstack{SmolLM3\\3B} & \shortstack{Tiny-Jev} & \shortstack{OpenJev\\0.6B} \\
\midrule
\multicolumn{14}{l}{\textbf{FANToM}} \\
Knowledge access & 1,446 & 81.3 & \textbf{82.6} & 77.4 & 82.0 & 63.0 & 70.3 & 60.7 & 77.6 & 49.3 & 49.1 & 78.4 & 32.2 \\
Beliefs & 227 & \textbf{73.1} & 35.2 & 24.2 & 21.6 & 39.6 & 38.3 & 28.2 & 17.6 & 32.2 & 4.4 & 12.3 & 7.0 \\
Control & 132 & 72.0 & 94.7 & \textbf{97.7} & 95.5 & 86.4 & 86.4 & 90.9 & 97.0 & 69.7 & 30.3 & 93.9 & 34.1 \\
All & 1,805 & \textbf{79.6} & 77.6 & 72.2 & 75.3 & 61.8 & 67.5 & 58.8 & 71.5 & 48.6 & 42.1 & 71.2 & 29.1 \\
\midrule
\multicolumn{14}{l}{\textbf{ToMATO}} \\
Knowledge access & 305 & 77.0 & 79.3 & \textbf{81.0} & 71.8 & 59.3 & 71.1 & 59.7 & 43.3 & 28.9 & 26.6 & 52.8 & 19.7 \\
Beliefs & 289 & \textbf{82.7} & 79.2 & 80.3 & 76.1 & 68.9 & 68.9 & 60.6 & 49.1 & 33.2 & 28.0 & 52.9 & 22.1 \\
Goals, desires and intentions & 546 & \textbf{86.1} & 84.8 & 84.2 & 77.8 & 68.1 & 73.8 & 65.8 & 54.4 & 35.0 & 39.2 & 55.3 & 17.2 \\
Emotions & 256 & \textbf{87.1} & 80.1 & 79.7 & 72.3 & 70.7 & 71.5 & 61.7 & 50.4 & 31.6 & 44.9 & 55.1 & 23.8 \\
All & 1,396 & \textbf{83.6} & 81.6 & 81.9 & 75.1 & 66.8 & 71.8 & 62.6 & 50.1 & 32.7 & 35.2 & 54.2 & 20.0 \\
\midrule
\multicolumn{14}{l}{\textbf{CogToM}} \\
Knowledge access & 212 & \textbf{66.5} & 63.7 & 66.0 & 58.0 & 60.4 & 45.8 & 42.0 & 52.8 & 34.4 & 46.7 & 36.8 & 42.5 \\
Beliefs & 440 & \textbf{76.6} & 72.7 & 70.7 & 63.4 & 72.7 & 56.1 & 58.6 & 35.7 & 39.5 & 35.0 & 30.5 & 28.6 \\
Goals, desires and intentions & 469 & \textbf{94.5} & 91.3 & \textbf{94.5} & 84.9 & 89.8 & 78.0 & 84.2 & 67.8 & 50.7 & 58.6 & 54.8 & 54.4 \\
Emotions & 331 & 92.4 & 89.1 & 90.9 & 84.3 & \textbf{92.7} & 86.4 & 81.0 & 66.8 & 45.6 & 71.3 & 56.2 & 58.0 \\
Social pragmatics & 197 & 94.4 & 90.9 & 91.9 & 90.4 & \textbf{95.9} & 83.8 & 79.7 & 51.3 & 42.1 & 46.2 & 53.3 & 57.4 \\
Unassigned & 505 & 67.1 & 64.6 & \textbf{75.2} & 59.2 & 61.2 & 53.9 & 46.7 & 30.7 & 25.7 & 42.8 & 19.6 & 27.7 \\
All & 2,154 & 81.3 & 78.1 & \textbf{81.5} & 72.2 & 77.7 & 66.5 & 65.1 & 49.4 & 39.4 & 49.7 & 39.9 & 42.5 \\
\midrule
\multicolumn{14}{l}{\textbf{Hi-ToM}} \\
Beliefs & 124 & 50.0 & 50.0 & \textbf{60.5} & 37.1 & 37.1 & 33.1 & 17.7 & 38.7 & 33.9 & 33.9 & 33.9 & 30.6 \\
Control & 30 & \textbf{100.0} & 80.0 & 83.3 & 66.7 & 60.0 & 43.3 & 20.0 & 43.3 & 30.0 & 40.0 & 16.7 & 36.7 \\
All & 154 & 59.7 & 55.8 & \textbf{64.9} & 42.9 & 41.6 & 35.1 & 18.2 & 39.6 & 33.1 & 35.1 & 30.5 & 31.8 \\
\midrule
\multicolumn{14}{l}{\textbf{ToMi}} \\
Beliefs & 96 & 50.0 & \textbf{83.3} & 67.7 & 44.8 & 31.2 & 45.8 & 49.0 & 46.9 & 75.0 & 19.8 & 62.5 & 37.5 \\
Control & 158 & 81.6 & 65.8 & 72.8 & 81.0 & 75.9 & 74.1 & 72.8 & \textbf{87.3} & 47.5 & 79.7 & 38.0 & 63.9 \\
All & 254 & 69.7 & \textbf{72.4} & 70.9 & 67.3 & 59.1 & 63.4 & 63.8 & 72.0 & 57.9 & 57.1 & 47.2 & 53.9 \\
\bottomrule
\end{tabular*}
\caption{Capability exact match (\%), part 2, using the same model order and scoring as Table~\ref{tab:current-capabilities-1}. Missing source--capability combinations are omitted.}
\label{tab:current-capabilities-2}
\end{table*}

\begin{table*}[!p]
\centering\small
\setlength{\tabcolsep}{4pt}
\begin{tabular*}{\textwidth}{@{\extracolsep{\fill}}lrr@{}}
\toprule
Model & All questions & Control-free \\
\midrule
GPT-3.5-1106 & 71.62 & 70.87 \\
Jev-1.13 & 73.21 & 69.08 \\
GPT-5.6-Luna & 70.85 & 68.55 \\
GPT-4o-mini & 66.29 & 61.59 \\
DeepSeek-V4-Flash & 65.40 & 60.22 \\
GPT-4.1-nano & 61.31 & 57.14 \\
Llama-3.1-8B & 57.64 & 54.03 \\
Open-Jev-2B & 56.60 & 51.64 \\
Valen & 49.53 & 49.47 \\
Tiny-Jev & 45.93 & 45.92 \\
SmolLM3-3B & 48.70 & 42.65 \\
OpenJev-0.6B & 43.19 & 38.86 \\
\bottomrule
\end{tabular*}
\caption{Benchmark-equal exact match with all 11,117 questions and after removing 920 native controls. Control-free scores retain 10,197 questions, including unmapped items. Rows descend by control-free accuracy.}
\label{tab:control_sensitivity}
\end{table*}

\begin{table*}[!p]
\centering\small
\setlength{\tabcolsep}{6pt}
\begin{tabular}{lrrr}
\toprule
Source & Questions & Kept & Wrong \\
\midrule
BigToM & 1,200 & 973 & 44 \\
OpenToM & 2,533 & 1,583 & 459 \\
ToMBench & 730 & 397 & 39 \\
SimpleToM & 891 & 418 & 146 \\
FANToM & 1,805 & 923 & 67 \\
ToMATO & 1,396 & 651 & 12 \\
CogToM & 2,154 & 1,311 & 57 \\
Hi-ToM & 154 & 33 & 0 \\
ToMi & 254 & 87 & 2 \\
Pooled & 11,117 & 6,376 & 826 \\
\bottomrule
\end{tabular}
\caption{Counts for Jev selection at probability $\geq0.90$. Coverage uses all questions, while retained error uses Kept. Native controls remain included.}
\label{tab:risk}
\end{table*}

\begin{table*}[!p]
\centering
\includegraphics[width=\textwidth]{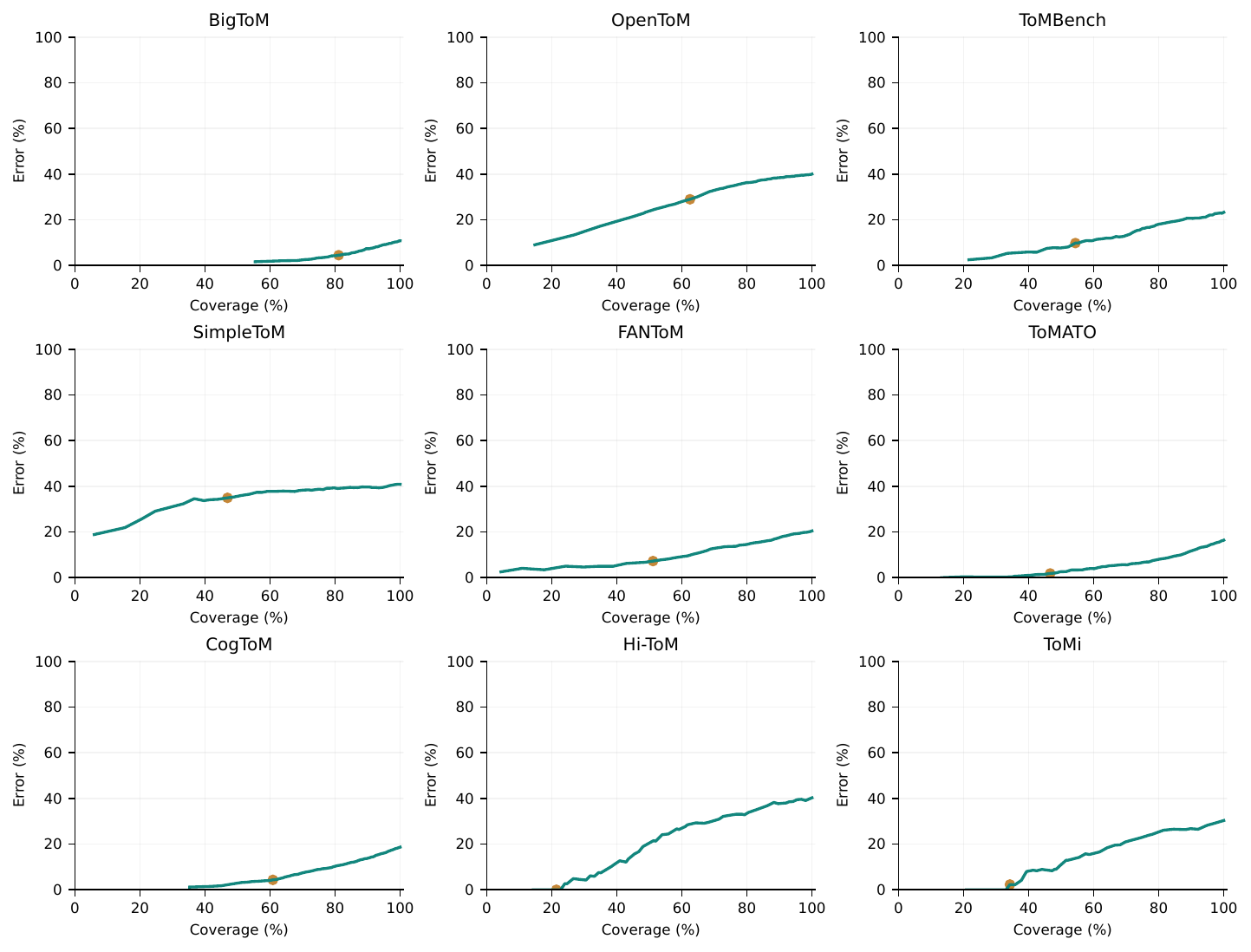}
\captionof{figure}{Jev risk--coverage curves on the same 11,117 primary questions as the main confidence analysis. Coverage is the percentage of all benchmark questions retained, and error is the percentage of retained answers that are incorrect. Each point retains all answers at or above one observed selected-answer probability, keeping ties together. Orange markers denote the 0.90 threshold.}
\label{fig:appendix_jev_risk_coverage}
\end{table*}

\begin{table*}[!p]
\centering\small
\setlength{\tabcolsep}{5pt}
\begin{tabular*}{\textwidth}{@{\extracolsep{\fill}}lr@{}}
\toprule
Annotation coverage / follow-up outcome & Questions \\
\midrule
Initial sample / targeted follow-up / overlap & 450 / 249 / 19 \\
Combined unique questions & 680 \\
Majority reference / unanimous / no majority & 219 / 211 / 30 \\
Majority references matching benchmark categories & 219/219 \\
\bottomrule
\end{tabular*}
\par\medskip
\begin{tabular*}{\textwidth}{@{\extracolsep{\fill}}lrrr@{}}
\toprule
Condition & Accuracy (\%) & $\Delta$ vs.\ no reasons (pp) & 95\% CI for $\Delta$ \\
\midrule
\multicolumn{4}{@{}l}{\textbf{Full group, original references}: 249 questions, 202 source families} \\
\midrule
No reasons & 96.39 & --- & --- \\
Aligned reasons & 99.00 & $+2.61$ & $[0.40, 5.00]$ \\
Swapped reasons & 25.30 & $-71.08$ & $[-75.41, -66.87]$ \\
\midrule
\multicolumn{4}{@{}l}{\textbf{Human-majority subset}: 219 questions, 180 source families} \\
\midrule
No reasons & 98.17 & --- & --- \\
Aligned reasons & 100.00 & $+1.83$ & $[0.47, 3.56]$ \\
Swapped reasons & 24.89 & $-73.29$ & $[-77.62, -68.92]$ \\
\bottomrule
\end{tabular*}
\caption{Annotation coverage and Jev intervention results under benchmark and human-majority references. The 19 overlaps count once. Majority references require at least two matching single-category votes, with unanimous cases included. Both orders yield 498 full-group and 438 majority-subset responses per condition. Accuracy is pooled (\%), differences are paired percentage points, and intervals use 2,000 within-benchmark family-bootstrap draws.}
\label{tab:human_annotation_coverage}
\label{tab:human_majority_intervention}
\end{table*}

\begin{table*}[!p]
\centering\small
\setlength{\tabcolsep}{4pt}
\begin{tabular*}{\textwidth}{@{\extracolsep{\fill}}llrrr@{}}
\toprule
Model & Materials & No reasons & Aligned & Crossed \\
\midrule
Jev-1.13 & Reused & 22.93 & 74.35 & 65.10 \\
Jev-1.13 & New & 88.10 & 91.79 & 87.70 \\
GPT-5.6-Luna & Reused & 36.31 & 70.05 & 67.19 \\
GPT-5.6-Luna & New & 81.79 & 88.90 & 86.15 \\
GPT-3.5-1106 & Reused & 36.36 & 72.35 & 60.52 \\
GPT-3.5-1106 & New & 82.21 & 88.38 & 81.58 \\
DeepSeek-V4-Flash & Reused & 30.67 & 60.10 & 55.83 \\
DeepSeek-V4-Flash & New & 66.87 & 80.95 & 77.49 \\
GPT-4o-mini & Reused & 30.04 & 60.27 & 47.08 \\
GPT-4o-mini & New & 78.77 & 87.19 & 74.37 \\
Open-Jev-2B & Reused & 32.91 & 51.52 & 37.23 \\
Open-Jev-2B & New & 63.49 & 71.77 & 65.32 \\
GPT-4.1-nano & Reused & 33.33 & 51.90 & 39.99 \\
GPT-4.1-nano & New & 67.89 & 76.58 & 59.19 \\
Llama-3.1-8B & Reused & 32.66 & 50.58 & 41.44 \\
Llama-3.1-8B & New & 58.62 & 68.58 & 53.44 \\
\bottomrule
\end{tabular*}
\par\medskip
\begin{tabular*}{\textwidth}{@{\extracolsep{\fill}}lrrr@{}}
\toprule
Jev, new materials & Repair (pp) & Damage (pp) & Net change (pp) \\
\midrule
Aligned & 6.96 & -3.27 & +3.69 \\
Crossed & 6.69 & -7.08 & -0.40 \\
\bottomrule
\end{tabular*}
\caption{Absolute answer-component accuracy on reused/new reason materials and Jev repair/damage contributions on new materials. Accuracies average benchmarks equally. Reused/new contain 1,803/2,958 existing questions. Repair and signed damage use all weighted outcomes.}
\label{tab:material_results_current}
\end{table*}

\begin{table*}[!p]
\centering\small
\setlength{\tabcolsep}{5pt}
\begin{tabular*}{\textwidth}{@{\extracolsep{\fill}}lrrrr@{}}
\toprule
Condition & Jev strict & Jev saved choice & GPT-3.5 strict & Jev invalid \\
\midrule
Answers only & 71.30 & 72.22 & 71.76 & 2 \\
Four original pairs & 85.65 & 85.65 & 79.63 & 0 \\
Four reassigned pairs & 92.13 & 92.59 & 91.20 & 1 \\
\shortstack[l]{Eight original-\\attachment pairs} & 81.02 & 84.26 & 76.39 & 10 \\
Eight reassigned pairs & 83.33 & 87.04 & 84.26 & 8 \\
All sixteen pairs & 76.39 & 82.41 & 62.50 & 16 \\
\bottomrule
\end{tabular*}
\caption{Matched-candidate accuracy (\%) on 108 questions in two orders, with equal benchmark weights. Jev invalid counts use 216 responses per row, and GPT-3.5 has no invalid outcomes. Answer-only measures answer accuracy, while pair rows measure joint accuracy. Post-hoc saved-choice scoring does not replace strict results.}
\label{tab:candidate_validity_review}
\end{table*}

\begin{table*}[!p]
\centering\small
\setlength{\tabcolsep}{5pt}
\begin{tabular}{lrr}
\toprule
Final failure type & Jev & DeepSeek \\
\midrule
Probability sum & 448 & 0 \\
Selected choice below maximum & 3 & 0 \\
Parsing / candidate ID & 0 & 4 \\
HTTP 429 & 0 & 843 \\
\midrule
Invalid / failed & 451 & 847 \\
All crossed responses & 9,522 & 9,522 \\
\bottomrule
\end{tabular}
\caption{Final crossed-pair failures. Counts are pooled across two orders, while Figure~\ref{fig:joint_diagnostic} uses equal benchmark weights. Earlier failed transport attempts are not additional outcomes.}
\label{tab:failure_summary}
\end{table*}

\begin{table*}[!p]
\centering
\begin{minipage}[t]{.49\textwidth}
\vspace{0pt}
\centering
\includegraphics[width=\linewidth]{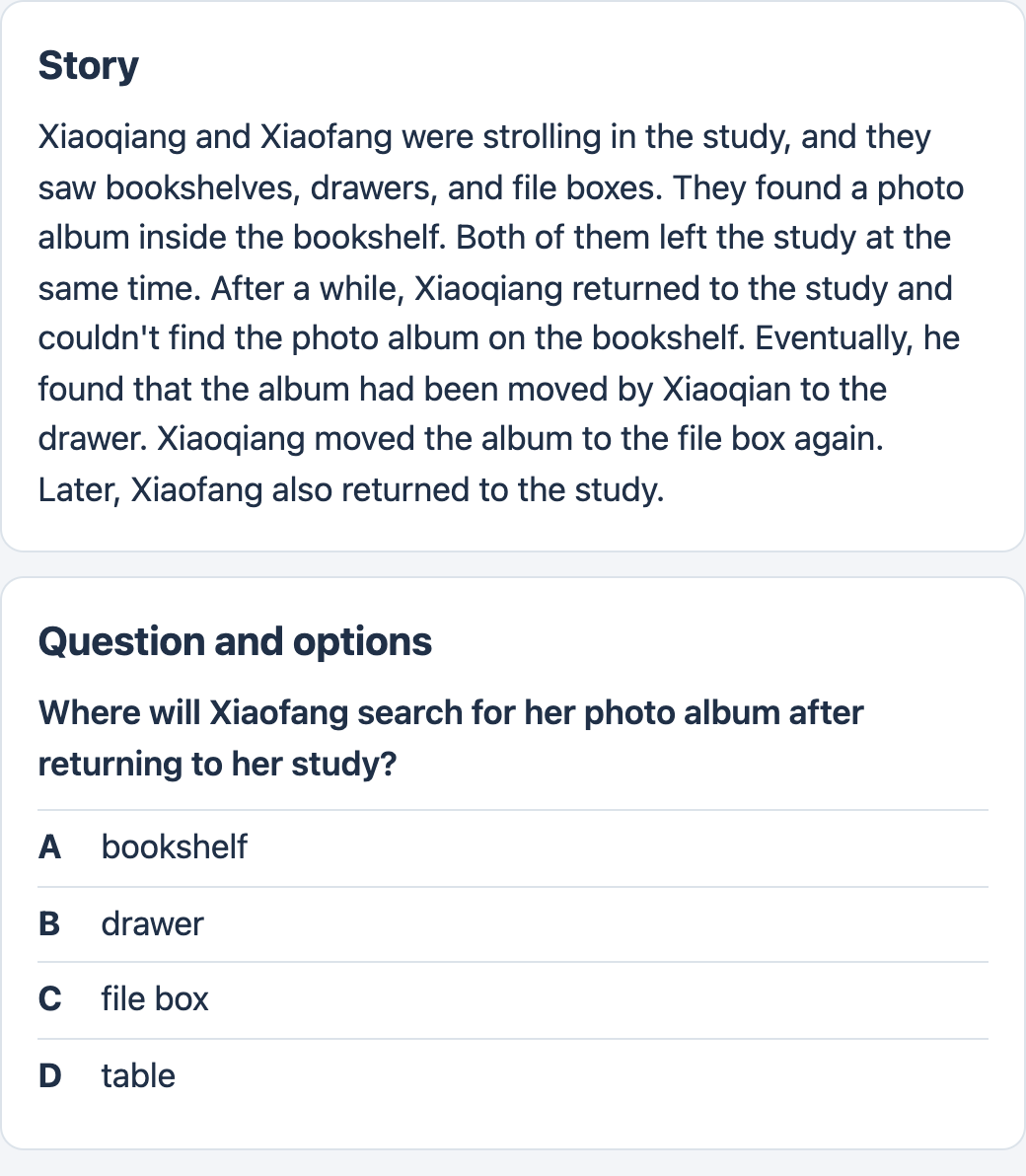}
\par\smallskip
\includegraphics[width=\linewidth]{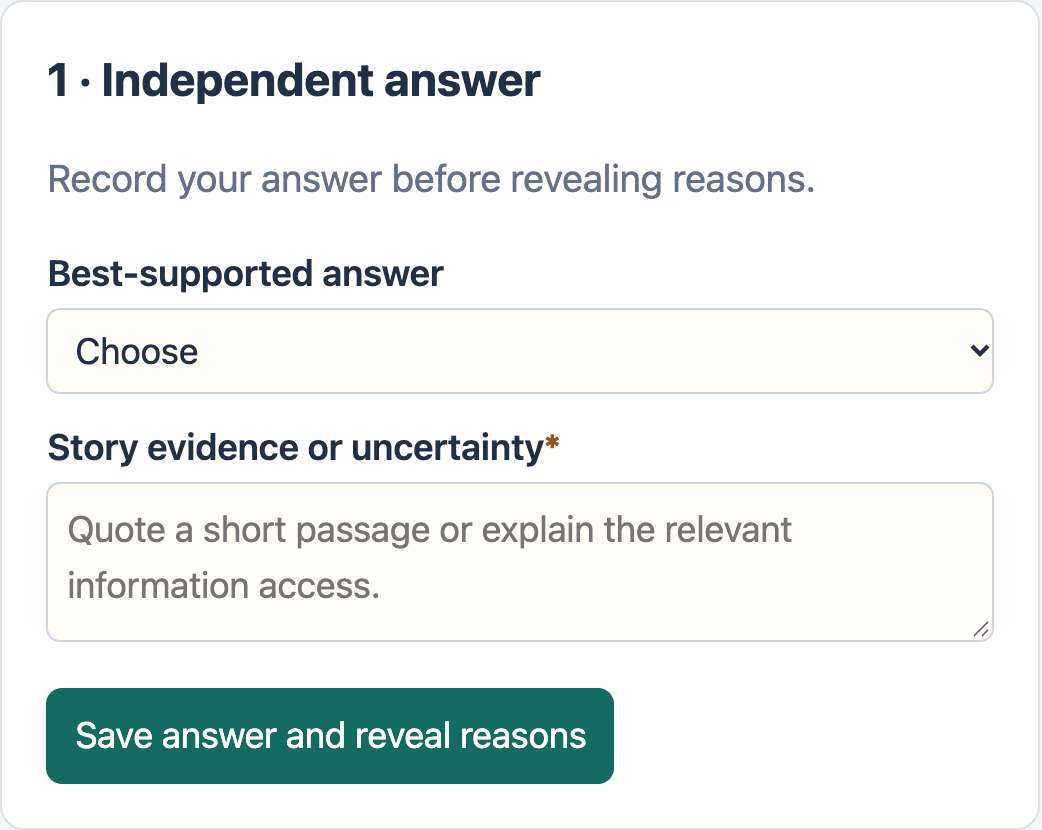}
\par\smallskip\small Initial Answer
\end{minipage}\hfill
\begin{minipage}[t]{.49\textwidth}
\vspace{0pt}
\centering
\includegraphics[width=\linewidth]{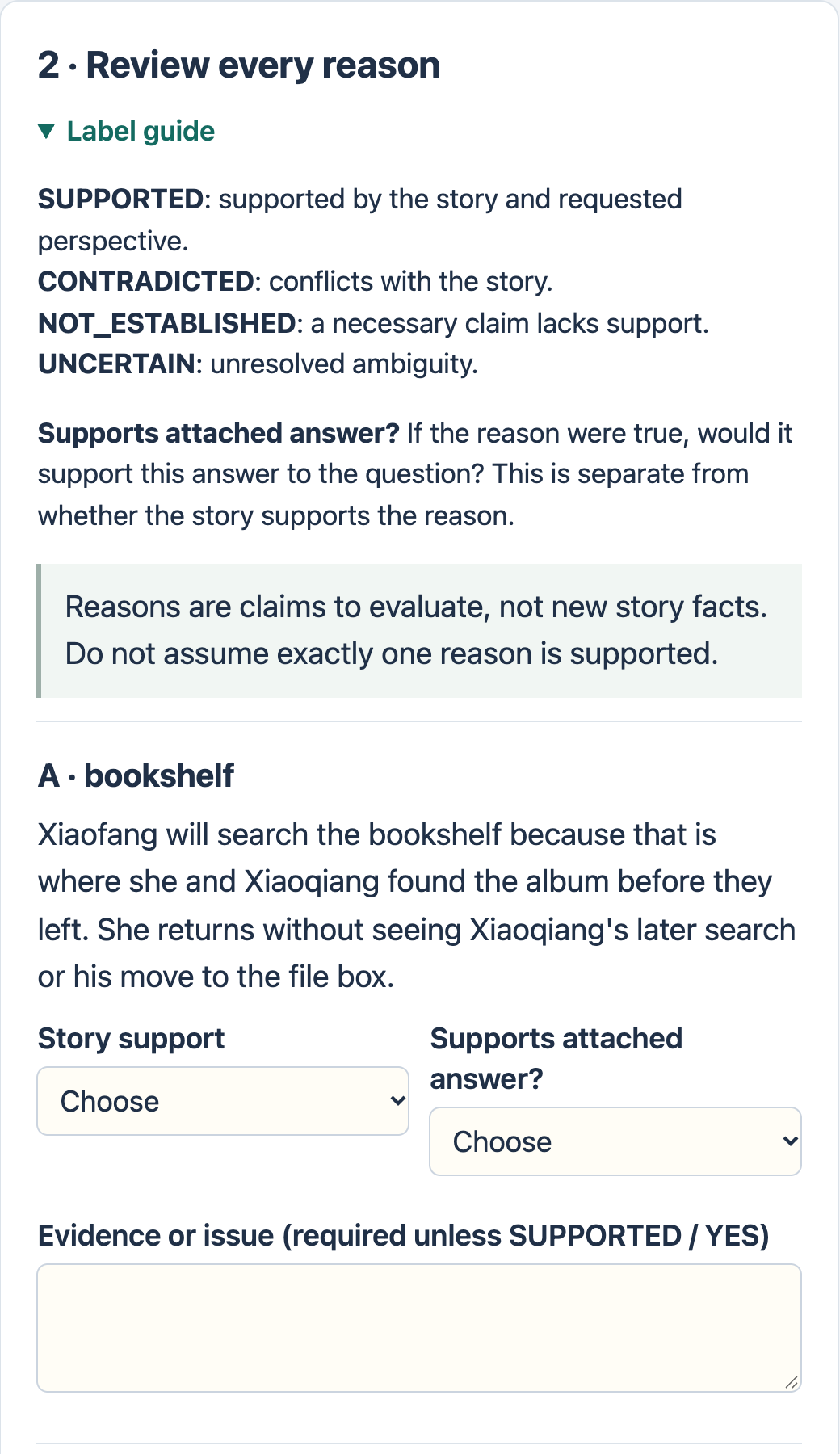}
\par\smallskip\small Reason Judgments
\end{minipage}
\captionof{figure}{Unfilled excerpts from the annotation interface. Initial Answer: story, question, options, and the initial-answer form, saved before reasons are revealed. Reason Judgments: label guide and the first option's reason-judgment fields, repeated for every option. The bilingual interface is shown in English for readability.}
\label{fig:human_annotation_interface}
\end{table*}

\begin{table*}[!p]
\centering\small
\setlength{\tabcolsep}{3pt}
\begin{tabular*}{\textwidth}{@{\extracolsep{\fill}}llrrrrr@{}}
\toprule
Model & Task & Belief/awareness & Original & Question & Predicted & Flipped \\
\midrule
Jev-1.13 & BigToM action & 84.50 & 92.00 & 98.00 & 97.50 & 97.00 \\
Jev-1.13 & SimpleToM behavior & 91.00 & 65.00 & 80.50 & 88.50 & 75.50 \\
Open-Jev-2B & BigToM action & 58.50 & 73.00 & 71.50 & 67.50 & 66.50 \\
Open-Jev-2B & SimpleToM behavior & 19.00 & 60.00 & 56.50 & 62.00 & 57.00 \\
Valen & BigToM action & 56.50 & 60.50 & 58.00 & 58.00 & 57.50 \\
Valen & SimpleToM behavior & 61.50 & 65.00 & 63.00 & 61.50 & 62.50 \\
Tiny-Jev & BigToM action & 62.50 & 55.00 & 55.50 & 54.50 & 58.00 \\
Tiny-Jev & SimpleToM behavior & 3.00 & 42.00 & 49.50 & 48.50 & 47.00 \\
OpenJev-0.6B & BigToM action & 54.50 & 53.50 & 51.00 & 51.00 & 50.50 \\
OpenJev-0.6B & SimpleToM behavior & 6.00 & 69.50 & 59.00 & 51.50 & 51.50 \\
\bottomrule
\end{tabular*}
\caption{Mental-state feedback accuracy (\%) by model and source, with 200 native upstream questions and 200 action/behavior targets per source. The upstream belief/awareness score is measured once per model. The four downstream conditions use the same target questions.}
\label{tab:feedback_extension_tasks}
\end{table*}

\begin{table*}[!p]
\centering\small
\setlength{\tabcolsep}{3pt}
\begin{tabular*}{\textwidth}{@{\extracolsep{\fill}}lrrrr@{}}
\toprule
Model & Question & Predicted & Fixed No & \shortstack{Changed\\choices} \\
\midrule
Jev-1.13 & 161 & 177 & 177 & 0 \\
Open-Jev-2B & 113 & 124 & 113 & 11 \\
Valen & 126 & 123 & 125 & 2 \\
Tiny-Jev & 99 & 97 & 94 & 5 \\
OpenJev-0.6B & 118 & 103 & 103 & 18 \\
\bottomrule
\end{tabular*}
\caption{Correct answers out of 200 SimpleToM behavior targets for question-only cues, predicted feedback, and the same-template fixed-No replay. Changed choices compares fixed No with predicted feedback on all 200 targets. Every output is valid. The replay selects the existing predicted or flipped response whose supplied answer is No, retaining the original attribution header and including every target exactly once.}
\label{tab:appendix_fixed_no}
\end{table*}

\begin{table*}[!p]
\centering\small
\setlength{\tabcolsep}{5pt}
\textbf{Reference prices and costs}\par\smallskip
\begin{tabular*}{\textwidth}{@{\extracolsep{\fill}}lrrrr@{}}
\toprule
Model & Input & Output & Total & Per 1k \\
\midrule
TypeSafe Jev & 0.0420 & 0.0000 & 0.3284 & 0.0295 \\
GPT-4.1 nano & 0.1000 & 0.4000 & 0.4876 & 0.0439 \\
GPT-4o mini & 0.1500 & 0.6000 & 0.7517 & 0.0676 \\
GPT-5.6 Luna & 0.2000 & 1.2000 & 9.9929 & 0.8989 \\
Llama-3.1-8B & 0.0500 & 0.0800 & 0.2463 & 0.0222 \\
DeepSeek-V4-Flash & 0.0108 & 1.2800 & 0.1535 & 0.0138 \\
GPT-3.5-1106 & 1.0000 & 2.0000 & 4.9278 & 0.4433 \\
\bottomrule
\end{tabular*}
\par\medskip
\textbf{Accuracy, observed time, and serving settings}\par\smallskip
\begin{tabular*}{\textwidth}{@{\extracolsep{\fill}}lrrp{.31\textwidth}rr@{}}
\toprule
Model & EM (\%) & Mean (ms) & Route / hardware & Workers & Attempts \\
\midrule
Jev-1.13 & 73.03 & 282.31 & OpenRouter, hardware NR & 1 & 1 \\
GPT-3.5-1106 & 71.62 & 1389.13 & OpenRouter, hardware NR & 8 & 3 \\
GPT-5.6-Luna & 70.85 & 7998.70 & OpenAI isolated session, hardware NR & 12 & 2 \\
GPT-4o-mini & 66.29 & 1591.08 & OpenRouter, hardware NR & 8 & 3 \\
DeepSeek-V4-Flash & 65.40 & 1188.86 & OpenRouter/DeepInfra, hardware NR & 8 & 3 \\
GPT-4.1-nano & 61.31 & 1753.78 & OpenRouter, hardware NR & 8 & 3 \\
Llama-3.1-8B & 57.64 & 652.05 & OpenRouter/DeepInfra, hardware NR & 8 & 3 \\
Open-Jev-2B & 56.60 & 464.34 & Local M4 Pro, 48 GiB & 1 & 1 \\
\bottomrule
\end{tabular*}
\caption{Resource comparison on 11,117 questions per deployment. The cost panel contains only seven priced models. Input/output tariffs are US dollars per million tokens, Total covers all questions, and Per 1k includes missing-usage allowances. GPT-3.5 uses a historical tariff. The timing panel includes all eight deployments. Attempts is the maximum total per question, and NR means not recorded.}
\label{tab:openrouter_prices}
\label{tab:resource_timing_current}
\label{tab:review_deployment_settings}
\end{table*}
\clearpage
\section{Prompts}
\label{app:prompts}
The following blocks preserve the recorded instruction text. They are ordered by the primary evaluation, reason intervention, material construction, and feedback study.

\subsection{Primary evaluation and output example}
Generative requests use the primary system instruction and user fields \texttt{Context:}, \texttt{Question:}, and \texttt{Options:}, retaining the frozen source order. The example illustrates the shared content and selected-answer probability.
\phantomsection\label{prompt:example}
\begin{promptbox}{Example input and output}
\textbf{Context:} Lisa has severe blisters on her feet. Lisa's dance class practices a new, intense choreography for an upcoming performance.\par
\textbf{Question:} What will Lisa's dance class likely do?\par
\textbf{c0:} advise Lisa to sit out and rest her feet\par
\textbf{c1:} encourage Lisa to keep up with the fast-paced routine\par\smallskip
\textbf{Typed output:}\par
\texttt{\{"choice": "c0", "probabilities":}\par
\texttt{\quad\{"c0": 0.9, "c1": 0.1\}\}}\par
\textbf{Generative label format:} A (corresponding to c0).
\end{promptbox}

\phantomsection\label{prompt:primary}
\begin{promptbox}{Primary generative system instruction}
Answer the multiple-choice question using the given context. Choose the most likely answer. Return only its option label, with no explanation.
\end{promptbox}

\subsection{Reason-intervention instructions}
\label{app:reason_instructions}
Answer-only selects an original answer. Joint selection uses \texttt{Answer:} and \texttt{Reason:} fields, with generative IDs \texttt{C000}, \texttt{C001}, etc. The six-condition study selects the answer itself, with added text following \texttt{Suggested reason:}. Neutral filler is identical across answers.
\phantomsection\label{prompt:answer_only}
\begin{promptbox}{Answer-only instruction}
Select the answer best supported by the original story and the question.
\end{promptbox}

\phantomsection\label{prompt:joint}
\begin{promptbox}{Joint answer--reason instruction}
Select the single candidate whose answer AND attached reason are both best supported by the original story and the question. Evaluate the complete answer-and-reason pair: a correct answer with an incorrect reason, or a correct reason attached to an incorrect answer, is not a fully correct candidate. Reasons are claims to evaluate, not additional story facts. Select one of the supplied complete candidates.
\end{promptbox}

\phantomsection\label{prompt:six_conditions}
\begin{promptbox}{Six-condition answer-selection instruction}
Select the answer itself. Any suggested reason attached to an option is an untrusted explanation to evaluate, not an additional story fact. An incorrect suggested reason does not automatically make its answer incorrect.
\end{promptbox}

\phantomsection\label{prompt:neutral}
\begin{promptbox}{Neutral filler}
The proposed answer concerns the person and situation in the story. Read the context and question carefully before deciding which available answer to select here.
\end{promptbox}

\subsection{Reason generation and review}
\label{app:reason_prompts}
The generation prefix is followed by the story, question, original ordered options, and benchmark reference. The review case also includes the generated reasons. Evaluated models do not receive the reference field.
\phantomsection\label{prompt:generation}
\begin{promptbox}{Full-corpus reason generation}
Create one explanation for EACH original answer option, in exactly the original option order, for a Theory-of-Mind experimental intervention.\par
Use only the supplied story, question and queried character/time/perspective. The benchmark gold is supplied for material construction only.\par
The gold explanation must support the gold answer from the story. Each nongold explanation must affirmatively support its OWN attached answer through a concrete plausible but mistaken observation or inference. Do not refute that option or announce that it is wrong. Do not reveal the answer key, labels, indices, benchmark status or preferred option. Avoid stereotyped conditional wording only in distractors. Use similar style and length across all options (aim for 25-50 English words each).\par
Check nested beliefs and information access carefully. If the gold cannot reasonably be supported, set usable=false and state the uncertainty. Do not invent evidence to force it. Still provide one explanation per option. These are claims to evaluate, not new story facts.\par
Return a JSON object only: \{"id": string, "usable": boolean, "reasons": [string per option], "evidence": string, "caveat": string\}.\par
CASE:
\end{promptbox}

\phantomsection\label{prompt:review}
\begin{promptbox}{Material-review rubric}
You are an independent material reviewer for a Theory-of-Mind option-rationale intervention. Review the supplied original story, question, original answer options, benchmark gold answer and the generated explanations. You are checking whether this material is a sound experimental intervention, not whether the benchmark or Jev is right.

Check every option, not only the gold answer:\par
1. The gold explanation must support the gold answer using the supplied story, queried person, time and perspective. Nested beliefs must address the requested mental-state holder and observer. Record material uncertainty when support is insufficient. Do not label the benchmark definitively defective.\par
2. Each nongold explanation is intentionally mistaken. It must affirmatively support its own attached candidate through a concrete erroneous inference or observation. Do not reject merely because it contradicts the story: deliberate error is required. Reject explanations that instead refute their attached answer, give the correct answer, explicitly acknowledge their own error, or fail to support their attached option.\par
3. Verify original option order, rationale count and answer mapping. Reject correctness leakage, an obvious label cue, or material that cannot implement the intended comparison.\par
4. Record minor length or style imbalance as warnings. Do not reject for a small deviation from the suggested 25--50 English words alone. Exclude only a concrete validity defect or unresolved material uncertainty. Do not repair, rewrite, or replace the original artifact.\par
5. Do not consult Jev outcomes, prior review judgments, historical success or the material generator's identity. All judgments must use the case below. This is gold-informed machine review, not human annotation or a measurement of Jev's internal reasoning.

Return JSON with id, approved (boolean), issues (list of concrete rejection reasons), warnings (list of residual limitations). A rejection must have an issue. An approval must have no unresolved validity issue. Review the full case before deciding.
\end{promptbox}

\subsection{Mental-state feedback suffixes}
\label{app:feedback_prompts}
Suffixes follow two newline characters. Newlines separate the question label, question, attribution, and selected answer text. Predicted and flipped conditions differ only in which upstream answer text is supplied, without references or probabilities.
\phantomsection\label{prompt:question_only_suffix}
\begin{promptbox}{Question-only feedback suffix}
Related mental-state question:\\
\texttt{\{upstream\_question\}}
\end{promptbox}

\phantomsection\label{prompt:state_suffix}
\begin{promptbox}{Predicted-state and flipped-state suffix}
Related mental-state question:\\
\texttt{\{upstream\_question\}}\\
Earlier answer selected by this model (may be incorrect):\\
\texttt{\{upstream\_selected\_option\_text\}}
\end{promptbox}

\end{document}